%% file: DeepMesoD_arXiv.tex
\documentclass[11pt]{article}

\usepackage[margin=1in]{geometry}
\usepackage{amsmath}
\usepackage{amssymb}
\usepackage{amsthm}
\usepackage{authblk}
\usepackage{booktabs}
\usepackage{float}
\usepackage{graphicx}
\usepackage{hyperref}
\usepackage{cleveref}
\usepackage{ifthen}
\usepackage[numbers,sort&compress,round]{natbib}
\usepackage{bibunits}
\usepackage{jabbrv}
\usepackage{physics}
\usepackage{placeins}
\usepackage{subcaption}
\usepackage{xcolor}
\usepackage{xparse}
\usepackage{xspace}

\newcommand{\method}{DeepMesoD\xspace}

\newboolean{shortarticle}
\setboolean{shortarticle}{false}
\newboolean{singlecolumn}
\setboolean{singlecolumn}{true}

\makeatletter
\let\arxiv@abstract\@empty
\let\arxiv@methods\@empty
\let\arxiv@acknowledgments\@empty
\let\arxiv@equal\@empty
\let\arxiv@corresponding\@empty

\RenewDocumentEnvironment{abstract}{+b}{\gdef\arxiv@abstract{#1}}{}
\newcommand{\significancestatement}[1]{}
\newcommand{\keywords}[1]{}
\newcommand{\leadauthor}[1]{}
\newcommand{\authorcontributions}[1]{}
\newcommand{\authordeclaration}[1]{}
\newcommand{\equalauthors}[1]{\gdef\arxiv@equal{#1}}
\newcommand{\correspondingauthor}[1]{\gdef\arxiv@corresponding{#1}}
\newcommand{\dates}[1]{}
\newcommand{\doi}[1]{}
\newcommand{\Firstpage}{}
\newcommand{\Endparasplit}{}
\let\ps@firststyle\ps@plain

\let\arxiv@maketitle\maketitle
\renewcommand{\maketitle}{
  \arxiv@maketitle
  \begingroup
    \renewcommand{\thefootnote}{\fnsymbol{footnote}}
    \renewcommand{\textsuperscript}[1]{}
    \footnotetext[1]{\arxiv@equal}
    \footnotetext[2]{\arxiv@corresponding}
  \endgroup
  \begin{center}
    \begin{minipage}{0.92\textwidth}
      \small
      \noindent\textbf{Abstract. }\arxiv@abstract
    \end{minipage}
  \end{center}
  \bigskip
}

\long\def\matmethods#1{\long\gdef\arxiv@methods{#1}}
\newcommand{\showmatmethods}{
  \section*{Materials and Methods}
  \arxiv@methods
}
\long\def\acknow#1{\long\gdef\arxiv@acknowledgments{#1}}
\newcommand{\showacknow}{
  \section*{Acknowledgments}
  \arxiv@acknowledgments
}
\makeatother

\ExplSyntaxOn
\NewDocumentCommand{\arxivAuthorNote}{m}{
  \tl_set:Nn \l_tmpa_tl {#1}
  \tl_replace_all:Nnn \l_tmpa_tl {1} {*}
  \tl_replace_all:Nnn \l_tmpa_tl {2} {\ensuremath{\dagger}}
  \textsuperscript{\normalfont\tl_use:N \l_tmpa_tl}
}
\ExplSyntaxOff
\makeatletter
\renewcommand{\AB@authnote}[1]{\protect\arxivAuthorNote{#1}}
\renewcommand{\@biblabel}[1]{#1.}
\makeatother

\begin{document}

\input{sections/00_frontmatter}
\clearpage
\input{sections/01_introduction}
\input{sections/02_framework}
\input{sections/03_ising_validation}
\input{sections/04_atomistic_systems}

\input{sections/05_discussion}
\input{sections/06_materials_and_methods}

\showmatmethods{}

\acknow{This work was supported by the National University of Singapore and the AI for Science Institute, Beijing, through the AISI--NUS Joint Research Initiative Fund 2025 Award. This work was also supported by the Ministry of Education (Singapore) through the Research Centre of Excellence program (grant EDUN C-33-18-279-V12, I-FIM).}
\showacknow

\bibliographystyle{pnas-new}
\bibliography{DeepMesoD}

\clearpage
\section*{Supplementary Information}
\setcounter{section}{0}
\setcounter{equation}{0}
\setcounter{figure}{0}
\setcounter{table}{0}
\renewcommand{\thesubsection}{\Alph{subsection}}
\renewcommand{\thesubsubsection}{\thesubsection.\arabic{subsubsection}}
\renewcommand{\theequation}{S\arabic{equation}}
\renewcommand{\thefigure}{S\arabic{figure}}
\renewcommand{\thetable}{S\arabic{table}}
\renewcommand{\theHsection}{supplement.\arabic{section}}
\renewcommand{\theHequation}{supplement.\arabic{equation}}
\renewcommand{\theHfigure}{supplement.\arabic{figure}}
\renewcommand{\theHtable}{supplement.\arabic{table}}
\graphicspath{{SI/}{./}}
\begin{bibunit}[pnas-new]
\makeatletter
\let\arxiv@SIlinkstart\hyper@natlinkstart
\let\arxiv@SIlinkbreak\hyper@natlinkbreak
\let\arxiv@SIanchorstart\hyper@natanchorstart
\def\hyper@natlinkstart#1{\arxiv@SIlinkstart{supplement.#1}}
\def\hyper@natlinkbreak#1#2{\arxiv@SIlinkbreak{#1}{supplement.#2}}
\def\hyper@natanchorstart#1{\arxiv@SIanchorstart{supplement.#1}}
\makeatother
\input{SI/sections/01_theory_and_methodology}
\input{SI/sections/02_ising}
\input{SI/sections/03_atomistic_spin}
\input{SI/sections/04_crbr3}
\FloatBarrier
\renewcommand{\refname}{Supplementary References}
\putbib[DeepMesoD]
\end{bibunit}

\end{document}

%% file: sections/00_frontmatter.tex
\title{Deep Mesoscopic Dynamics}

\author[a,1]{Yue Zhao}
\author[b,1]{Shiqi Wu}
\author[a,c,d]{Yihang Ouyang}
\author[a]{Sergey Grebenchuk}
\author[a]{Magdalena Grzeszczyk}
\author[e,f]{Xiaoxu Zhao}
\author[a,c,2]{Kostya S. Novoselov}
\author[a,c,2]{Maciej Koperski}
\author[a,b,2]{Qianxiao Li}

\affil[a]{Institute for Functional Intelligent Materials, National University of Singapore, Singapore 117544, Singapore}
\affil[b]{Department of Mathematics, National University of Singapore, Singapore 117543, Singapore}
\affil[c]{Department of Materials Science and Engineering, National University of Singapore, Singapore 117575, Singapore}
\affil[d]{Integrative Sciences and Engineering Programme, NUS Graduate School, National University of Singapore, 119077, Singapore}
\affil[e]{School of Materials Science and Engineering, Peking University, Beijing 100871, China}
\affil[f]{AI for Science Institute, Beijing 100871, China}

\leadauthor{Zhao}

\significancestatement{
Domains and interfaces control the properties of complex material systems,
but simulating their evolution poses a dilemma:
Microscopic models that resolve individual degrees of freedom are too costly to run over the scale at which patterns form, whereas bulk averages erase these patterns entirely.
Deep Mesoscopic Dynamics learns the missing intermediate-scale description directly from short microscopic trajectories in a thermodynamically consistent way,
resolving effective free energy densities and spatial interaction operators.
We validate the approach on a spin system where the intermediate description
is analytically available, and then apply it to study magnetic materials without
known mesoscopic equations.
The learned mesoscopic models run orders of magnitude faster than the simulations that trained them and extend to larger regions without retraining.
}

\authorcontributions{Y.Z., S.W., K.S.N., M.K., and Q.L. designed the research. Y.Z. and S.W. performed numerical simulations. Y.O., S.G., and M.G. performed experiments. K.S.N., M.K., and Q.L. supervised the research. All authors contributed to writing and revising the paper.}
\authordeclaration{No competing interest.}
\equalauthors{\textsuperscript{1}Equal contribution.}
\correspondingauthor{\textsuperscript{2}Corresponding authors: \mbox{kostya@nus.edu.sg}; \mbox{msemaci@nus.edu.sg}; \mbox{qianxiao@nus.edu.sg}.}

\keywords{mesoscopic dynamics $|$ fluctuation--dissipation theorem $|$ coarse-graining $|$ thermodynamic structure $|$ scientific machine learning}

\begin{abstract}
Microscopic simulations resolve spatial patterns but are feasible only over small domains and short timescales, whereas bulk descriptions erase those patterns through spatial averaging.
Mesoscopic fields preserve this spatial organisation, but their effective equations are known only for selected microscopic systems and scaling limits.
We introduce Deep Mesoscopic Dynamics (\method{}), which uses neural networks to learn an effective mesoscopic field equation from microscopic trajectories at a prescribed coarse-graining scale. For the nonconserved dissipative dynamics studied here, \method{} learns the free-energy functional with a spatial interaction kernel and a state-dependent dissipation operator constrained by the fluctuation--dissipation relation.
We first validate the construction in the Ising--Kac model, where the learned mesoscopic equation recovers the known hydrodynamic limit and rollouts remain stable far beyond the training horizon.
We then treat two systems without closed-form mesoscopic descriptions.
For a two-dimensional Heisenberg ferromagnet governed by atomistic spin dynamics, \method{} learns a mesoscopic model whose predictions match atomistic references up to $1\,\mathrm{ns}$, and a rollout with an unseen initial condition remains bounded up to $25\,\mathrm{ns}$.
For layered CrBr$_3$, the learned kernel captures the suppression and sign reversal of interlayer coupling at a stacking fault whose location and strength are not supplied as inputs.
Controlled changes to the learned interaction show how fault number and placement shape multistep magnetic responses, complementing our magneto-photoluminescence measurements.
Across the two atomistic systems, \method{} achieves a three-to-four-order-of-magnitude reduction in wall-clock time compared with atomistic spin dynamics.
\end{abstract}

\dates{Edited by Editor Name; This manuscript was compiled on \today}
\doi{\url{www.pnas.org/cgi/doi/10.1073/pnas.XXXXXXXXXX}}

\maketitle
\thispagestyle{firststyle}
\ifthenelse{\boolean{shortarticle}}{\ifthenelse{\boolean{singlecolumn}}{\abscontentformatted}{\abscontent}}{}

\Firstpage

%% file: sections/01_introduction.tex
\noindent
In magnets, alloys, and other many-body systems, microscopic interactions organise spins or atoms into spatial patterns whose evolution controls material behaviour. 
For example, magnetic domains can store information~\cite{hubert1998magnetic}, with localised spin textures such as skyrmions explored for spintronic and probabilistic-computing applications~\cite{grebenchuk_topological_2024,zazvorka2019thermal}, while antiphase domains and phase boundaries shape the evolution of alloy microstructures~\cite{AllenCahn1979,cahn1958free}. 
Predicting how these structures form, move, and interact requires a dynamical description that resolves their spatial variation. 
Microscopic simulations such as atomistic spin dynamics and molecular dynamics evolve the coupled dynamics of individual spins or atoms, but their computational cost limits the spatial region over which complete patterns can be resolved~\cite{e2011principles,jia2020pushing,guo2022extending}. 
On the other hand, descriptions based only on bulk averages such as total magnetisation or mean composition cannot capture domains, textures, or interfaces.

A mesoscopic description lies between these two levels and provides an efficient means to simulate these spatial features at realistic timescales.
This description is captured by a field $\phi\colon\Omega\to\mathbb R^q$ defined on a spatial domain $\Omega$, and $q$ is the number of field components.
The field value $\phi(\mathbf{x})$ at each spatial location $\mathbf{x} \in \Omega$ is obtained by a local coarse-graining map acting on the microscopic configurations near $\mathbf{x}$.
For example, for magnetic systems $\phi(\mathbf{x})$ may be the average magnetic moment near $\mathbf{x}$.
\Endparasplit
\noindent
The mesoscopic field $\phi$ describes the formation, motion,  and interaction of  spatial structures without the need to record microscopic arrangement below the prescribed scale (Figure~\ref{fig:pipeline}, left).
In particular, if an accurate evolution equation can be derived for $\phi$,
simulations can then be realised over spatial and temporal scales that are relevant to probe important phenomena,
such as domain formation and phase transitions.
However, even when the microscopic dynamics and coarse-graining map are known,
determining an accurate evolution equation for $\phi$ is a central open problem in dynamics modelling.

Classically, such mesoscopic field dynamics are commonly constructed in two ways.
The first prescribes phenomenological forms for the free energy and field dynamics using symmetry, conservation laws, and empirical knowledge.
For example, Landau-Ginzburg theory~\cite{landau1937theory,landau1980statistical} posits that
$\phi$ evolves according to a gradient flow with respect to a phenomenological free energy functional
\[
\mathcal F[\phi]
=
\int_\Omega
\left[
V\bigl(\phi(\mathbf{x})\bigr)
+
\frac{\kappa}{2}
\left|\nabla\phi(\mathbf{x})\right|^2
\right]
d\mathbf{x}.
\]
where the local free energy density $V = a_2\phi^2/2+a_4\phi^4/4$, 
 $\kappa>0$, $a_2\in\mathbb R$ and $a_4>0$, yielding the Allen-Cahn equation~\cite{AllenCahn1979,cahn1958free}
 $\partial_t \phi = - \delta \mathcal{F} / \delta \phi$.
The coefficients $a_2$, $a_4$, $\kappa$ can be fitted, but the model remains restricted to the prescribed form of energy,
and is not an accurate description of any particular dissipative dynamical process at the mesoscopic scale.

The second approach directly derives a mesoscopic field equation from a microscopic dynamical model. For selected microscopic systems, a controlled scaling limit yields an effective hydrodynamics equation for the coarse field~\cite{KipnisLandim1999,Spohn1991}. The Ising--Kac model with Glauber dynamics is one example~\cite{presutti2009scaling,de1994glauber}. For more complex systems, eliminating microscopic variables that are not represented in the coarse field can produce conditional expectations, memory terms, or high-dimensional integrals that are not available in a computable form~\cite{zwanzig2001nonequilibrium}. In practice, dynamic mean-field models obtain a computable equation by prescribing a mean-field free energy and selected field dynamics~\cite{fraaije1997dynamic}. The resulting equation then depends on these modelling approximations, which can omit effects that remain relevant at the chosen coarse-graining scale.

To address these difficulties, we introduce \method, which constructs accurate mesoscopic field dynamics without the need
for phenomenological prescriptions or complex derivations.
Instead, we consider a highly general class of dissipative dynamics with unknown dissipation operators and free energy functionals,
where the thermal noise and the dissipation operator satisfy a fluctuation-dissipation relation.
These operators are parametrised by structured neural networks and trained directly on microscopic observations to
obtain accurate mesoscopic descriptions that can then be used to simulate and analyse the system on large temporal and spatial scales.

We first validate \method{} in the Ising--Kac model, whose hydrodynamic limit provides analytic references for the learned field equation~\cite{presutti2009scaling,de1994glauber}. We then apply the same construction to a Cr\textsubscript{2}Ge\textsubscript{2}Te\textsubscript{6}-like two-dimensional ferromagnet~\cite{Skubic2008,evans2014atomistic} and layered CrBr$_3$~\cite{Grzeszczyk2023ExcitonMagnetization,Kipczak2024ResonantRamanCrBr3,grebenchuk_correlations_2025,Fullerton2026ChiralitySwitchingSkyrmion,yang_controlling_2023,yao_multiple_2023}, two systems whose mesoscopic equations are not known in closed form. Atomistic-spin trajectories provide microscopic references in both systems. In CrBr$_3$, the modelled multistep response is consistent with our magneto-photoluminescence measurements. Controlled numerical experiments with the learned stacking-fault interaction show how fault number and placement affect the calculated field response.

%% file: sections/02_framework.tex
\section*{DeepMesoD Methodology}
\label{sec:methodology}

\begin{figure*}[t]
  \centering
  \includegraphics[width=\textwidth]{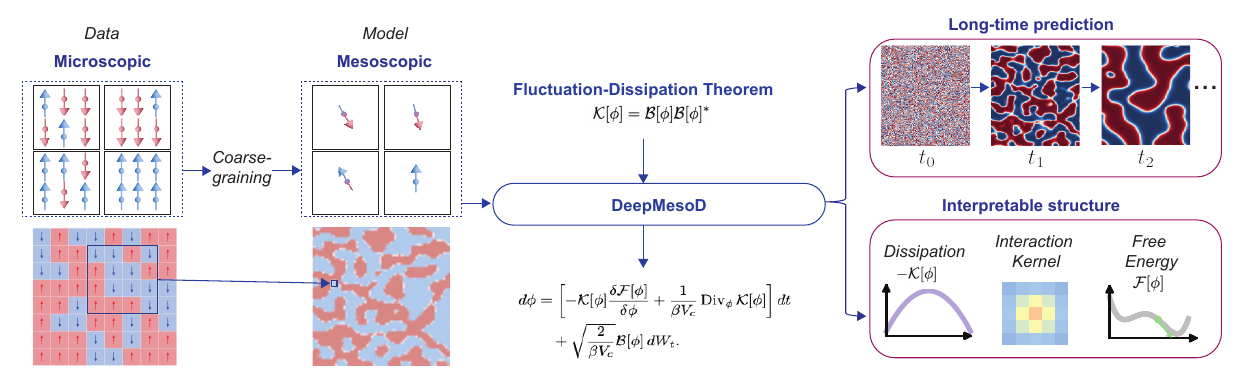}
  \caption{\textbf{Workflow of \method.} Microscopic trajectories are coarse-grained into observations of a mesoscopic field at a prescribed spatial scale. \method{} uses these trajectories to estimate the physical structures in a mesoscopic field equation within a prescribed dynamical class.}
  \label{fig:pipeline}
\end{figure*}

\paragraph*{From microscopic trajectories to field observations}
We first describe how microscopic trajectories provide coarse-grained field observations.
While our framework is designed for general dissipative mesoscopic dynamics, for clarity we specialise our discussion to magnetic spin systems defined on a lattice.
We denote the index set of microscopic spins by $\mathcal I$ and the state space of each spin by $\mathsf S$.
The microscopic configuration $\sigma(t)=(\sigma_j(t))_{j\in\mathcal I}\in\mathsf S^{\mathcal I}$ is assumed to evolve as a continuous-time Markov process.
For example, in the Ising--Kac ferromagnet, $\mathsf S=\{-1,+1\}$ and $\sigma(t)$ evolves as a Markov jump process under Glauber dynamics~\cite{ising1925beitrag,glauber1963time,presutti2009scaling}.
Another example is classical Heisenberg magnets, where each atomic spin is represented by a three-dimensional unit vector.
In this case, $\mathsf S=\mathbb S^2$ is the unit sphere, and $\sigma_j(t)$ specifies the orientation of atomic spin $j$.
The trajectories are recorded at discrete observation times.
At each recorded time, we compute coarse observations at $N$ spatial locations $\mathbf x_i\in\Omega$.
For each location, $\mathcal C(\mathbf x_i)\subset\mathcal I$ contains the microscopic spins in its averaging neighbourhood, with $V_c=|\mathcal C(\mathbf x_i)|$ fixed within each system.
The observation map $\mathcal Q$ gives the value $z_i(t)$ at that position:
\[
z_i(t)=\mathcal Q(\mathbf x_i;\sigma(t))
=\frac{1}{V_c}\sum_{j\in\mathcal C(\mathbf x_i)}\sigma_j(t).
\]
These averages give scalar magnetisation for Ising--Kac ($q=1$) and three-component magnetisation for the atomistic spin systems ($q=3$), where each microscopic spin is a unit vector.
We seek a mesoscopic field $\phi\colon\Omega\times[0,\infty)\to\mathbb R^q$ whose dynamics reproduce the observations,
\[
z_i(t)\approx\phi(\mathbf x_i,t).
\]
Thus each training observation pairs a spatial position $\mathbf x_i$ with a field value $z_i(t)$.
We write $\Phi$ for the space of these fields at any fixed time.

\paragraph*{Field dynamics and fluctuation--dissipation structure}
We aim to learn an effective equation for the evolution of $\phi$ from the observations $z_i(t)$.
We consider fields whose spatial integral is not conserved and whose coarse-grained observations exhibit thermal fluctuations, motivating a description that combines relaxation and noise.
We model the relaxation as free-energy-driven dissipation, giving the stochastic field equation in It\^o form
\begin{equation}
\partial_t\phi
=-\mathcal K[\phi]\frac{\delta\mathcal F[\phi]}{\delta\phi}
+\frac{1}{\beta V_c}\operatorname{Div}_\phi\mathcal K[\phi]
+\sqrt{\frac{2}{\beta V_c}}\mathcal B[\phi]\xi,
\label{eq:spde}
\end{equation}
where $\beta=T^{-1}$ is the inverse temperature in Boltzmann units.
The noise $\xi$ is a zero-mean, $p$-component Gaussian field that is white in space and time, with
\[
\mathbb E[\xi(\mathbf x,t)\xi(\mathbf y,s)^{\mathsf T}]
=I_p\,\delta(\mathbf x-\mathbf y)\,\delta(t-s),
\]
where $I_p$ is the $p\times p$ identity matrix.

This hypothesis space is inspired by Landau's model, but with important distinctions.
First, the free-energy functional $\mathcal F$  is not assumed to involve a polynomial energy density, nor gradient-based domain penalisation ($\|\nabla\phi\|^2$).
Instead, we minimally constrain this functional to satisfy locality and translational invariance on homogeneous domains (details below).
Second, we explicitly model the rate of thermal dissipation via a general dissipation operator $\mathcal K$.
We similarly constrain $\mathcal K$ by symmetry principles, in the spirit of Onsager's reciprocal relations~\cite{onsager1931reciprocal}.
Third, we explicitly model state-dependent thermal noise ($\mathcal B$), which is assumed to satisfy a fluctuation--dissipation relation $\mathcal K[\phi]=\mathcal B[\phi]\mathcal B[\phi]^*$, where $*$ denotes the adjoint operator.
The divergence term $(\beta V_c)^{-1}\operatorname{Div}_{\phi}\mathcal K[\phi]$ represents the thermal drift in the It\^o form of the stochastic differential equation with state-dependent dissipation. As the fields and operators are defined on function spaces, rigorous definitions of this term and the stochastic-integral formulation are given in the SI Appendix, section~1.
In summary, instead of guessing explicit function forms as in phenomenological modelling,
here we constrain the physical quantities by broad symmetry and invariance principles,
and leave exact functional and operator identification as a learning problem under these constraints.
Below we present the precise constraints we apply to each quantity.

\paragraph*{Free energy and spatial interaction}
We constrain the free-energy density at each position to depend on the local field and on a variation field that describes its spatial environment:
\begin{equation}
\mathcal F[\phi]
=\int_\Omega\rho\!\left(\phi(\mathbf x),(\mathcal L_J\phi)(\mathbf x)\right)\,d\mathbf x.
\label{eq:learned_free_energy}
\end{equation}
The energy density $\rho\colon\mathbb R^q\times\mathbb R^q\to\mathbb R$ depends on the local value $\phi(\mathbf x)$ and the variation field $\mathcal L_J\phi$.
The bounded operator $\mathcal L_J\colon\Phi\to\Phi$ is defined by
\[
(\mathcal L_J\phi)(\mathbf x)
=\int_\Omega J(\mathbf x,\mathbf y)[\phi(\mathbf x)-\phi(\mathbf y)]\,d\mathbf y,
\]
where $J\colon\Omega\times\Omega\to\mathbb R$ is a spatial interaction kernel learned from data.
We use the normalisation convention
\[
\int_\Omega J(\mathbf x,\mathbf y)\,d\mathbf y=1,
\qquad\mathbf x\in\Omega.
\]
Notice that the variation field $\mathcal L_J\phi$ vanishes by construction on spatially uniform $\phi$.
The energy-density function $\rho$ and the interaction kernel $J$ are learned jointly from the coarse-field trajectories.
This generalises the Landau form of the free energy functional, corresponding to the choice $\rho(u,v)=g(u)+\kappa\|v\|^2$, where $g$ is a local energy density and $J$ is chosen so the last term is a non-local approximation of the gradient norm, which describes interfacial energy.
Our parametrisation, however, allows for non-additive interactions between bulk and interfacial energies,
and this turns out to be important for the van der Waals magnets we investigate later.

\paragraph*{Factorised dissipation operator}
We now discuss the construction of the dissipation operator.
Since we enforce the fluctuation--dissipation relation through the factorisation $
\mathcal K=\mathcal B\mathcal B^*$, it suffices to parametrise the factor $\mathcal B$.
This factorisation makes $\mathcal K[\phi]$ self-adjoint and positive semidefinite, as expected for a dissipation operator.
At a fixed field state, $\mathcal B[\phi]$ maps a $p$-component input noise field to a perturbation of the $q$-component magnetisation field.
Its matrix-valued kernel $b[\phi](\mathbf x,\mathbf y)$ has $q$ rows indexing magnetisation components at $\mathbf x$ and $p$ columns indexing noise components at $\mathbf y$.
For an input field $\eta$, the factor acts as
\[
[\mathcal B[\phi]\eta](\mathbf x)
=\int_\Omega b[\phi](\mathbf x,\mathbf y)\eta(\mathbf y)\,d\mathbf y.
\]
In this work we study short-range interactions and assume a locality radius $R>0$.
The factor $b[\phi](\mathbf x,\mathbf y)$ vanishes when the distance between $\mathbf x$ and $\mathbf y$ exceeds $R$, and depends on $\phi$ only through field values within distance $R$ of $\mathbf x$.
For example, single-spin flips in Ising--Kac Glauber dynamics produce local instantaneous fluctuations in nonoverlapping block averages, while their rates depend on surrounding spins, motivating these locality assumptions.
On periodic homogeneous domains, sharing the same local coefficient rule across positions makes the factor translation equivariant and reusable on larger domains without retraining, at fixed coarse-graining scale and numerical resolution. 
Numerical details are given in Materials and Methods.

\paragraph*{Learning from field observations}
We learn the energy density $\rho$, the spatial interaction kernel $J$, and factor kernel $b$ from the observations $z_i(t)$ at positions $\mathbf x_i$. We use a multilayer perceptron (MLP) to represent $\rho$, trainable interaction coefficients to represent $J$ (here assumed to be piecewise constant, thus uniquely defined by a square matrix of function values), and a convolutional neural network (CNN) to represent the dependence of $b$ on local field values.
For training, we first learn the noise factor $\mathcal B$ from fluctuations in the observed field increments after subtracting their fitted mean drift.
The fluctuation--dissipation relation then gives $\mathcal K=\mathcal B\mathcal B^*$.
With the factor parameters fixed, we jointly fit $\rho$ and $J$ to the observed field increments using the complete deterministic drift in \eqref{eq:spde}.
After training, we numerically integrate the learned drift in \eqref{eq:spde} to simulate the mesoscopic field on a coarse grid. The learned equations enable simulations on much larger domains without retraining (SI Appendix, section~2G) and run thousands to tens of thousands of times faster than atomistic spin dynamics for the two magnetic materials (Materials and Methods).
Details are given in Materials and Methods and the SI Appendix.
For the magnetic systems below, we use the conventional notation $m$ for a scalar magnetisation field and $\mathbf m$ for a vector magnetisation field, in place of $\phi$.

%% file: sections/03_ising_validation.tex
\section*{Validation against hydrodynamic limit}

\paragraph*{Microscopic model and mesoscopic references}
The Ising model is a classical model of magnetic order and phase transitions built from lattice spins with two possible orientations~\cite{ising1925beitrag}. Its Kac variant couples spins over many lattice spacings through an interaction kernel (e.g. Gaussian) while keeping the total interaction strength fixed~\cite{presutti2009scaling}.
Under stochastic spin flips governed by Glauber dynamics~\cite{glauber1963time}, its known hydrodynamic limit provides a reference for validating the mesoscopic dynamics learned by \method{}~\cite{de1994glauber,presutti2009scaling}.

We consider   the  two-dimensional periodic   microscopic lattice $\mathcal I=(\mathbb Z/L\mathbb Z)^2$  of linear size $L$, with   spin configuration $\sigma(t)\in\{-1,+1\}^{\mathcal I}$.
A spin at site $j$ flips at the heat-bath Glauber rate
\[
\begin{aligned}
c_j(\sigma)=\frac{1}{2}\left[1-\sigma_j\tanh\!\left(\beta H_j(\sigma)\right)\right],\\
H_j(\sigma)=\sum_{\ell\ne j}J(j-\ell)\sigma_\ell+h_{\mathrm{ext}},
\end{aligned}
\]
where $H_j$  is  the   local interaction field and $h_{\mathrm{ext}}$  is the  applied field .
 We use a   normalised  periodic Gaussian Kac   kernel  of range $\epsilon_{\mathrm K}$ in   unit-domain coordinates on $\mathbb T^2$, the unit periodic domain. The kernel $J(j-\ell)$ denotes its corresponding  lattice weight .
  The flips satisfy detailed balance with respect to the Boltzmann distribution $\Pr(\sigma)\propto\exp[-\beta\mathcal H(\sigma)]$, with  Ising--Kac   Hamiltonian
\[
\mathcal H(\sigma)
=-\frac{1}{2}\sum_{\substack{j,\ell\in\mathcal I\\j\ne\ell}}J(j-\ell)\sigma_j\sigma_\ell
-h_{\mathrm{ext}}\sum_{j\in\mathcal I}\sigma_j.
\]
 Averaging $V_c$ spins   at  each coarse   position $\mathbf x_i$  gives the   scalar observations $z_i(t)\approx m(\mathbf x_i,t)$ .
The simulation procedure is given in Materials and Methods and the SI Appendix, section~2A.

  Under Kac scaling, each averaging cell contains increasingly many spins ($V_c\to\infty$) while its diameter becomes small relative to the interaction range. Microscopic  fluctuations average out  , and the hydrodynamic limit gives the deterministic magnetisation equation~\cite{de1994glauber,presutti2009scaling}
 \begin{equation}
\partial_t m(\mathbf x,t)
=-m(\mathbf x,t)+\tanh\!\bigl(\beta[(J*m)(\mathbf x,t)+h_{\mathrm{ext}}]\bigr),
\label{eq:ising_hydrodynamic}
\end{equation}
where $(J*m)(\mathbf x)=\int_{\mathbb T^2}J(\mathbf x-\mathbf y)m(\mathbf y)\,d\mathbf y$ and $\int_{\mathbb T^2}J(\mathbf r)\,d\mathbf r=1$.
We rewrite \eqref{eq:ising_hydrodynamic} in the deterministic form of \eqref{eq:spde}, with the reference dissipation operator $\mathcal K$ defined accordingly.
The reference free energy $\mathcal F_{\mathrm{ref}}$ has the form \eqref{eq:learned_free_energy}, with energy density
\[
\begin{aligned}
\rho(m,\ell)&=g_{\mathrm{ref}}(m)-h_{\mathrm{ext}}m+\tfrac12m\ell,\\
\ell&=\mathcal L_Jm,
\end{aligned}
\]
where $\ell=m-J*m$.
The homogeneous equilibrium free-energy density is
\[
g_{\mathrm{ref}}(m)
=\frac{1}{\beta}\left[
\frac{1+m}{2}\log\frac{1+m}{2}
+\frac{1-m}{2}\log\frac{1-m}{2}\right]
-\frac{1}{2}m^2+C.
\]
Here $C$ is an arbitrary additive constant. We consider two temperatures on either side of the critical temperature $T_c=1$ at zero field.
At $T=0.75$, spins collectively align to form magnetic domains, whereas at $T=1.5$, thermal fluctuations prevent spontaneous magnetic order.

\begin{figure*}[t]
  \centering
\includegraphics[width=\textwidth]{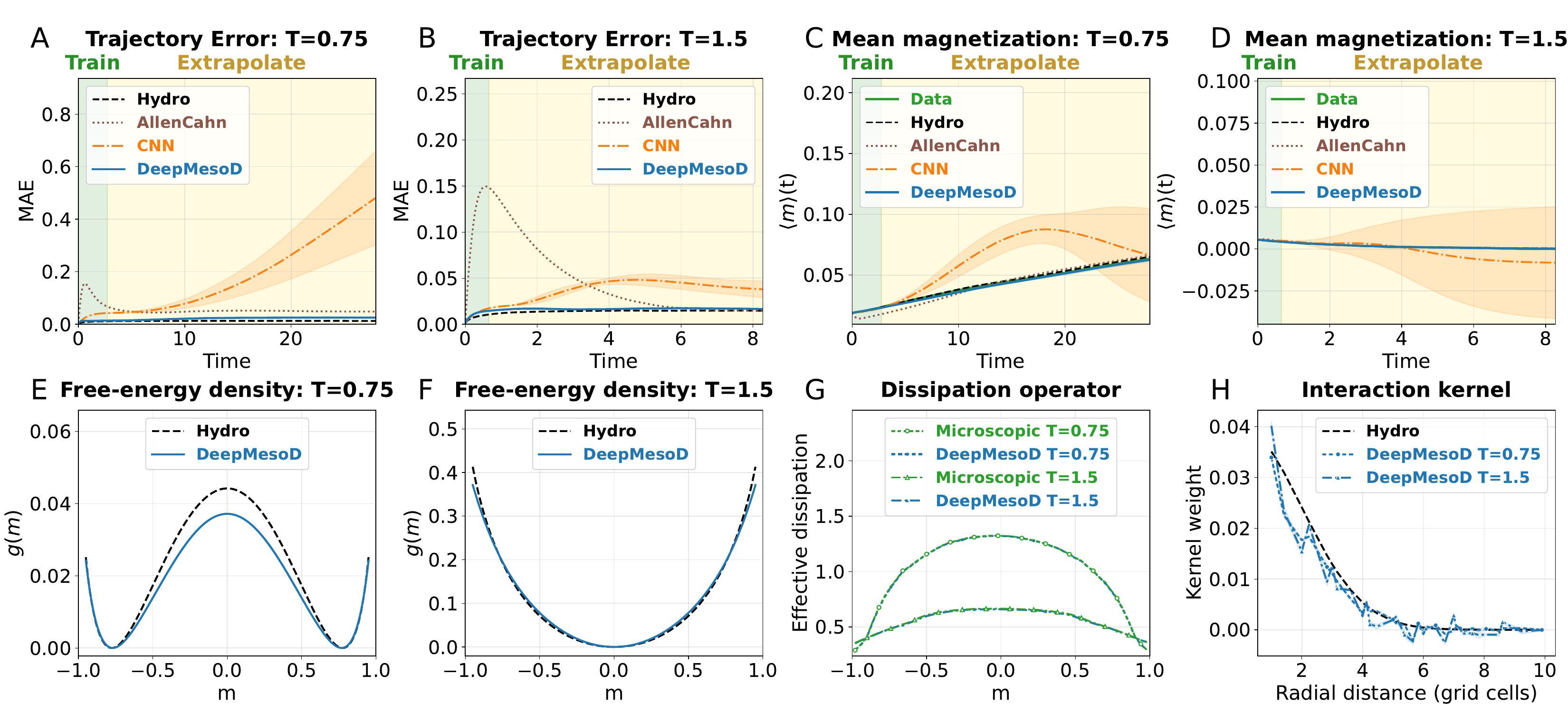}
	  \caption{{\textbf{Validation of \method{} in the Ising--Kac system with hydrodynamic limit.} The \method{} rollouts are noise-free with the It\^o thermal drift retained.
			  \textbf{A,B.} Field mean absolute error at each evaluation time for \(T=0.75\) and \(T=1.5\). Green shading marks the training interval, and yellow shading marks the subsequent rollout-only interval.
		  \textbf{C,D.} Mean magnetisation during the same rollouts.
\textbf{E,F.} Homogeneous free-energy density \(g(m)\), compared with the hydrodynamic-limit reference. Each curve is shifted independently so that its minimum is zero.
\textbf{G.} Conditional dissipation-operator response \(\mathbb E\{[\mathcal K[\phi]\mathbf 1](\mathbf x)\mid\phi(\mathbf x)=m\}\), compared with the reference in \eqref{eq:ising_cov_reference} computed from the corresponding microscopic trajectories.
\textbf{H.} Off-centre interaction weights, averaged over offsets with the same radial distance and compared with the Gaussian Ising--Kac kernel. For DeepMesoD and the CNN baseline, lines show the mean across three training seeds, and shaded bands indicate one standard deviation.}}
  \label{fig:ising_main}
\end{figure*}

The hydrodynamic limit describes deterministic evolution, whereas finite averaging cells  still exhibit fluctuations.
Spin-flip activity determines the noise covariance and hence a dissipation reference through the fluctuation--dissipation relation.
Let $\bar c_i=V_c^{-1}\sum_{j\in\mathcal C(\mathbf x_i)}c_j(\sigma)$  be the mean instantaneous flip activity in cell $i$.   Conditioning  on the   observation $z_i=m$ gives
 \begin{equation}
K  _{\mathrm{cov}} (m)=2\beta\,\mathbb E[\bar c_i\mid   z _i=m].
\label{eq:ising_cov_reference}
\end{equation}
We compare the learned dynamics with \eqref{eq:ising_hydrodynamic}, the  homogeneous energy density with $g_{\mathrm{ref}}$, and the dissipation response with $K_{\mathrm{cov}}$ (\Cref{fig:ising_main}).
The free-energy and finite-block covariance derivations are given in the SI Appendix, sections~2B--C.

\paragraph*{{Simulation and model setup}}
The simulations use $L=4096$ and $\epsilon_{\mathrm K}=1/64$. Each $32\times32$ block contains $V_c=1024$ spins, and the observations are stored on a $128\times128$ array. The microscopic initial conditions are sampled from local Bernoulli probabilities whose mean magnetisation is specified by multiscale Gaussian random fields. At $T=0.75$, the first 500 time steps from each of three independently generated trajectories are used for training. At $T=1.5$, the first 100 time steps from each trajectory are used for training. A separate initial field is generated for evaluation, and three microscopic realisations initialised from this field are averaged to form the reference trajectory. The evaluation rollouts contain 5,000 steps at $T=0.75$ and 1,000 at $T=1.5$. Each reported learned-model result at a given temperature is based on three independent training runs with different random seeds. The temporal sampling and training details are given in Materials and Methods and the SI Appendix, sections~2A and~2D.

For the Ising validation, we use $\rho(m,\ell)=g(m)-h_{\mathrm{ext}}m+\tfrac12m\ell$ and learn the homogeneous density $g$ and a symmetric interaction $\widehat J$, with $\ell=\mathcal L_{\widehat J}m$. The hat distinguishes the learned interaction from the reference $J$. This choice allows a direct comparison of $g$ with $g_{\mathrm{ref}}$ while learning the spatial interaction from data.

We compare \method{} with the hydrodynamic-limit equation \eqref{eq:ising_hydrodynamic}, a local Allen--Cahn approximation, and a patch CNN fitted directly to $\partial_t m$. The Allen--Cahn approximation replaces the nonlocal Kac interaction by its local second-order expansion, while the CNN provides an unconstrained data-driven baseline. Their numerical definitions are given in the SI Appendix, section~2E. An FNO trained on the same short Ising trajectories produced unstable autoregressive rollouts, so we omit its results from \Cref{fig:ising_main} and present an FNO comparison for the atomistic spin system below.

\paragraph*{{Validation of dynamics and physical structure}}
\Cref{fig:ising_main} evaluates the learned equation in the ordered regime at $T=0.75$ and in the disordered regime at $T=1.5$, both at $h_{\mathrm{ext}}=0$. \method{} maintains a low field mean absolute error over 5,000 evaluation steps at $T=0.75$ and 1,000 at $T=1.5$, while the CNN accumulates substantially larger error (panels \textit{A} and \textit{B}). The spatial mean magnetisation follows the microscopic and hydrodynamic-limit references throughout the same intervals (panels \textit{C} and \textit{D}). The Allen--Cahn approximation has its largest error early in the rollout and lower error later, consistent with its second-order expansion better approximating the nonlocal interaction as the field becomes smoother.

The same figure compares the learned constitutive functions with their references. The homogeneous density $g$ reproduces the double-well structure at $T=0.75$ and the single-well structure at $T=1.5$ and tracks the reference profiles $g_{\mathrm{ref}}$ (panels \textit{E} and \textit{F}). The conditional response of the learned dissipation operator, defined in the caption, agrees with the reference $K_{\rm cov}(m)$ defined in \eqref{eq:ising_cov_reference} and estimated from microscopic trajectories (panel \textit{G}). For panel \textit{H}, we average the learned off-centre interaction weights over offsets at the same distance from the centre. The resulting radial profile agrees with the Gaussian reference, although training does not constrain weights at equal distances to be equal.
Rollouts on larger periodic domains retain comparable errors at the same coarse-graining scale and numerical resolution (SI Appendix, section~2G).

%% file: sections/04_atomistic_systems.tex
\section*{Discovery of mesoscopic structure in atomistic systems}
\label{subsec:asd}

We next apply \method{} to atomistic spin dynamics trajectories from two magnetic systems whose mesoscopic equations are not known in closed form~\cite{Skubic2008,evans2014atomistic}. In a Cr\textsubscript{2}Ge\textsubscript{2}Te\textsubscript{6}-like two-dimensional ferromagnet, we examine whether the learned equation recovers the anisotropy, exchange, and state-dependent dissipation required for long rollouts~\cite{jenkins2022breaking}. In CrBr$_3$, we use the learned stacking-fault interaction in controlled numerical experiments that vary the number and placement of fault interfaces. The resulting responses reveal how this interaction organises multistep magnetic switching. Once learned, these neural-network field equations permit stable simulations over larger spatial domains at substantially lower computational cost than atomistic spin dynamics.

\paragraph*{Microscopic model and atomistic spin dynamics}
At the atomistic level, both systems are described by a classical Heisenberg
spin Hamiltonian, with a unit spin $\mathbf{s}_i\in S^2$ carrying moment
$\mu_s$ at each magnetic site. For these vector spins, we write $\mathbf s_i$ in place of the microscopic variable $\sigma_i$ used above. A general form of the Hamiltonian is
\[
\mathcal{H} = -\sum_{i<j} J_{ij}\,\mathbf{s}_i\cdot\mathbf{s}_j
    - k_u\sum_i (\mathbf{s}_i\cdot\mathbf{e})^2
    - \mu_s\sum_i \mathbf{s}_i\cdot\mathbf{B}
    + \mathcal{H}_{\mathrm{dip}},
\]
combining exchange with couplings $J_{ij}$, uniaxial anisotropy of strength
$k_u$ about the easy axis $\mathbf{e}$, Zeeman coupling to an applied field
$\mathbf{B}$, and the long-range dipolar interaction
$\mathcal{H}_{\mathrm{dip}}$. The Cr$_2$Ge$_2$Te$_6$-like ferromagnet and the
CrBr$_3$ film differ in lattice geometry and material parameters
$\{J_{ij}, k_u, \mu_s\}$, and the stacking fault in the latter enters as a local
modification of the exchange couplings across the fault plane.

Given this Hamiltonian, each spin evolves under the stochastic
Landau-Lifshitz-Gilbert (LLG) equation
\begin{equation}
\begin{split}
\frac{d\mathbf{s}_i}{dt}&=-\frac{\gamma}{1+\alpha^2}\,
\mathbf{s}_i\times\bigl(\mathbf{H}_i+\alpha\,\mathbf{s}_i\times\mathbf{H}_i\bigr),\\
\mathbf{H}_i&=-\frac{1}{\mu_s}\frac{\partial \mathcal{H}}{\partial\mathbf{s}_i}
    +\boldsymbol{\zeta}_i,
\end{split}
\end{equation}
which combines reversible precession about the effective field $\mathbf{H}_i$,
Gilbert damping with strength $\alpha$, and thermal noise $\boldsymbol{\zeta}_i$
whose covariance is fixed by the fluctuation--dissipation relation at
temperature $T$. We integrate these equations with
VAMPIRE v7.0.0~\cite{Skubic2008,evans2014atomistic}, which supplies the microscopic
trajectories that \method{} coarse-grains.

In both systems, the coarse observations are obtained by the same block average as before, now applied to the three-component spins:
\[
z_i(t)=V_c^{-1}\sum_{j\in\mathcal{C}(\mathbf{x}_i)}\mathbf{s}_j(t)\in\mathbb{R}^3.
\]
We learn a vector field $\mathbf m$ with $z_i(t)\approx\mathbf m(\mathbf x_i,t)$ and $|\mathbf m|\le 1$.
Unlike the Ising--Kac model, whose coarse-grained constitutive structure is analytically known, no general constitutive description is available for atomistic spin systems governed by the LLG equation.

\paragraph*{Cr\textsubscript{2}Ge\textsubscript{2}Te\textsubscript{6}-like two-dimensional van der Waals ferromagnet}
Two-dimensional van der Waals magnets such as CrI\textsubscript{3} and Cr\textsubscript{2}Ge\textsubscript{2}Te\textsubscript{6} provide atomically thin platforms for exploring low-dimensional magnetism and developing spintronic devices~\cite{burch2018magnetism}. Simulating two-dimensional magnets at experimentally relevant spatial and temporal scales is essential because their collective behaviour may not be captured in small computational domains. Large-scale atomistic simulations, for example, have revealed finite-size magnetic order beyond the conventional Mermin--Wagner limit, but at substantial computational cost, involving millions of spins and days of simulation~\cite{jenkins2022breaking}. By learning thermodynamically consistent mesoscopic dynamics from atomistic trajectories, \method{} enables cheaper spatial scale-up and long-horizon prediction beyond the reach of direct microscopic simulation.
It is also of scientific interest to evaluate if an Allen--Cahn-type phenomenological model
can quantitatively capture the mesoscopic dynamics of these magnets.

For the two-dimensional van der Waals ferromagnet, we specialise the general Hamiltonian above by restricting $J_{ij}$ to a uniform nearest-neighbour coupling $J$, setting the applied field to zero, and neglecting dipolar interactions.
With the easy axis chosen along $z$, the resulting anisotropic Heisenberg Hamiltonian reads
\[
H(\mathbf{s})=-\sum_{\langle i,j\rangle}J\,\mathbf{s}_i\cdot\mathbf{s}_j
-k_u\sum_i (s_i^z)^2,
\]
where each site carries a classical unit spin $\mathbf{s}_i\in S^2$, $J>0$ is the ferromagnetic exchange coupling, and $k_u>0$ is the uniaxial anisotropy strength.
The uniaxial anisotropy stabilises finite-temperature magnetic order, and the exchange and anisotropy parameters are adopted from Ref.~\cite{jenkins2022breaking}.
Under this effective modelling hypothesis, \method{} uses the same finite-scale,
thermodynamically consistent dissipative construction introduced in
\eqref{eq:spde} and applied to the Ising--Kac system.
The learned interaction kernel obeys the unit-sum constraint, $\sum_r\widehat J(r)=1$, and for the present vector-valued magnetisation field, the dissipation construction is extended from the scalar case by replacing each scalar coupling with a $3\times3$ block. Specifically, $b_i(s;\mathbf m)\in\mathbb R^{3\times3}$ acts on the three spin components of neighbouring coarse cells, while the factorisation $K=BB^T$ and the finite-range construction remain unchanged. The factor uses the same $3\times3$ neighbourhood of noise sources as in the Ising--Kac system.

\begin{figure}[!t]
\centering
\includegraphics[width=\textwidth]{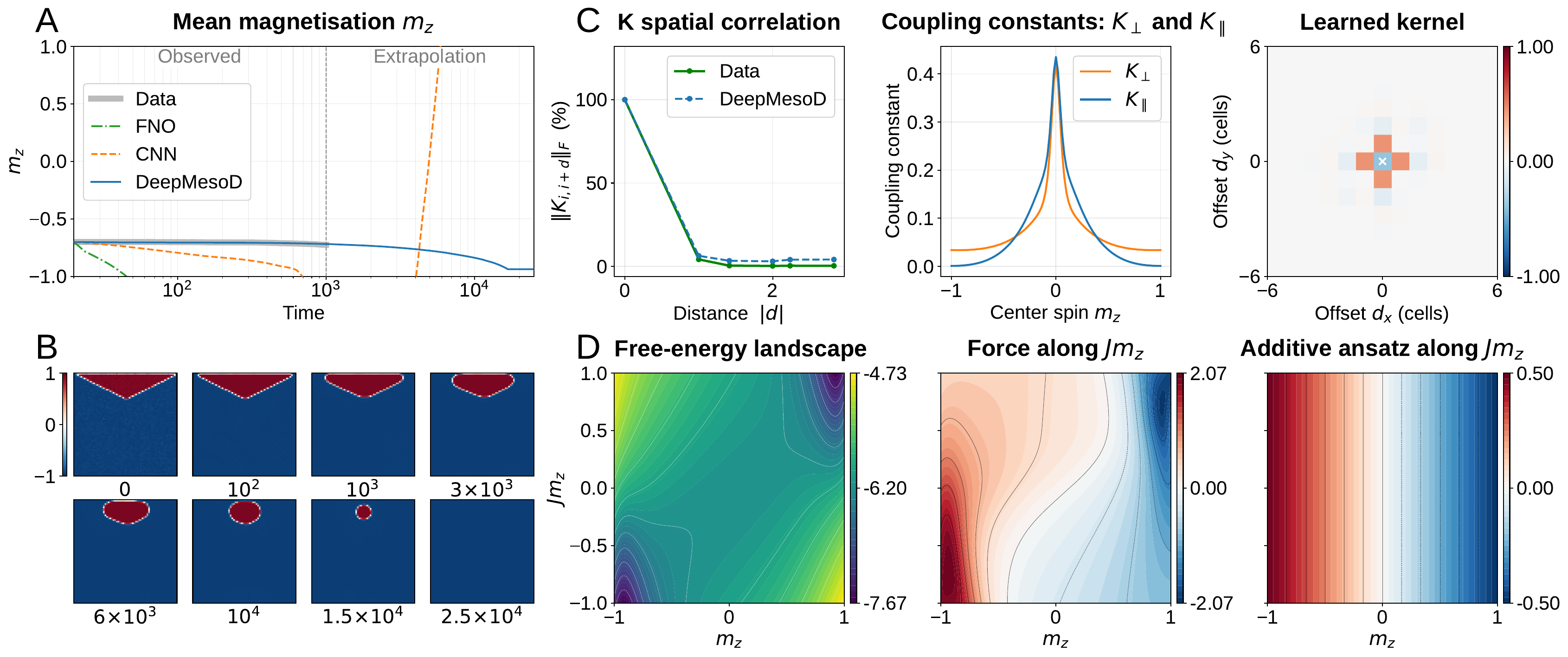}
\caption{\textbf{Long-term stability and thermodynamic-structure recovery in atomistic spin dynamics.}
    \textbf{A.} Long-term rollout from a triangular-domain initial condition on the same lattice. The \method{} prediction is noise-free with the It\^o thermal drift retained.
    Training uses random-initial-state trajectories. The triangular-domain initial condition is used for geometric-initial-condition extrapolation, not for fitting these stages.
    The dashed boundary at $1\,\mathrm{ns}$ separates the interval with an atomistic reference (``Observed'') from the interval without a reference (``Extrapolation''), ending at $25\,\mathrm{ns}$. Time labels are in ps. The atomistic reference permits accuracy comparisons over $0$--$1\,\mathrm{ns}$, but extending it to $25\,\mathrm{ns}$ is computationally costly. The $1$--$25\,\mathrm{ns}$ learned-model rollout assesses numerical stability and qualitative relaxation.
    \textbf{B.} Predicted magnetisation fields at the indicated times in ps. The triangular minority domain contracts towards a uniform magnetised state.
    \textbf{C.} Recovered Kernel.
    Left: spatial decay of the learned dissipation operator, quantified by the Frobenius norm of the block $K_{i,i+d}$ with distance $d$ relative to the on-site block.
    Middle: dissipation coupling constants $K_\perp$ and $K_\parallel$ as functions of centre spin $m_z$, exhibiting an even (symmetric) profile that peaks near the disordered state and reflecting the expected axial anisotropy between the two coupling modes.
    Right: learned interaction kernel, revealing a dominant nearest-neighbour structure consistent with short-range exchange coupling.
    \textbf{D.} Recovered free energy.
    Left: free-energy landscape $F(m_z, Jm_z)$ exhibiting a double-well profile with minima at $m_z=\pm1$, consistent with the underlying uniaxial anisotropy.
    Middle: thermodynamic force $\partial F/\partial(Jm_z)$ recovered by \method{}, which is a general nonlinear function of both $m_z$ and $Jm_z$ and exhibits non-trivial contour structure.
    Right: the hydrodynamic limit $\partial F/\partial(Jm_z)$, which assumes an additive dependence on $m_z$ alone, yields a purely striped pattern independent of $Jm_z$, demonstrating that the additive assumption fails to capture the full free-energy structure of this system.}
\label{fig:asd_composite}
\end{figure}

\method{} is trained on $10$ independent atomistic spin dynamics trajectories generated from random spin initialisations. Each trajectory is simulated for $250\,\mathrm{ps}$ with a microscopic time step of $0.1\,\mathrm{fs}$, and configurations are saved every $1\,\mathrm{ps}$. Training data are drawn from the first $150\,\mathrm{ps}$ of each trajectory, with the remaining $100\,\mathrm{ps}$ reserved for later-time evaluation. We further assess out-of-distribution generalisation with a triangular-domain initial condition, in which a single triangular magnetic domain is embedded in an otherwise uniformly magnetised background. Additional evaluations on domain-wall configurations are reported in the SI, section~3D.
Two unstructured baselines are compared: the patch-wise CNN used in the Ising--Kac experiments and a Fourier neural operator (FNO~\cite{li2020fourier}) that learns the field evolution as a black-box operator.
For the triangular-domain initial condition, atomistic reference data are available over $0$--$1\,\mathrm{ns}$, where prediction errors can be evaluated (\Cref{fig:asd_composite}A and SI Appendix, section~3D). The learned-model rollout extends to $25\,\mathrm{ns}$, a timescale beyond the practical reach of ASD for the same system under a comparable computational budget. Over $1$--$25\,\mathrm{ns}$, the predicted mean magnetisation remains bounded and approaches $-1$ as the minority domain shrinks and the system relaxes towards a uniformly magnetised state (\Cref{fig:asd_composite}B). In contrast, the CNN and FNO baselines become unstable during long rollouts. 

No violations of $|\mathbf{m}|\leq1$ were detected for \method{} in the reported tests. Table~\ref{tab:ood_violations} shows 0.00\% unphysical predictions across the three tested initial conditions, against ${>}37\%$ and ${>}81\%$ for CNN and FNO.

\begin{table}[htbp]
\centering
\caption{\textbf{Physical constraint violations in learned-model predictions.} Percentage of grid points where the predicted spin magnitude exceeds the unit sphere ($|\mathbf{m}|>1$) across three magnetic configurations, without any post-processing. The 0.00\% entries for \method{} indicate zero detected violations.}
\label{tab:ood_violations}
\begin{tabular}{lccc}
\toprule
\textbf{Model} & \textbf{Random} & \textbf{Domain Wall} & \textbf{Triangular domain} \\ \midrule
\method        & \textbf{0.00\%} & \textbf{0.00\%}      & \textbf{0.00\%}     \\
CNN            & 37.59\%         & 94.48\%              & 94.44\%             \\
FNO            & 81.66\%         & 96.66\%              & 98.31\%             \\ \bottomrule
\end{tabular}
\end{table}

Beyond stable rollout, \method reconstructs the free-energy functional, dissipation operator, and interaction kernel of the ASD system directly from data.
The learned free-energy landscape exhibits a double-well profile with minima at $m_z \approx \pm 1$, consistent with the underlying uniaxial anisotropy (\Cref{fig:asd_composite}D, left). These minima provide energetically favoured magnetised states, supporting a physical interpretation of the stable relaxation observed in the rollouts.
Compared with the additive dependence assumed in the hydrodynamic limit of the Ising--Kac model, the learned thermodynamic force $\partial F/\partial(Jm_z)$ exhibits a nonlinear dependence on both $m_z$ and $Jm_z$. The resulting two-dimensional response surface contains clear non-separable structure (\Cref{fig:asd_composite}D, middle and right), demonstrating that the general parametrisation of the free-energy density used in \method{} can represent thermodynamic couplings beyond an additive form.
The learned \mbox{dissipation} is likewise strongly localised in space. The Frobenius norm of the coupling block $K_{i, i+d}$ is largest on the on-site block ($d=0$) and drops sharply already at $d=1$, so the operator is effectively on-site dissipation(\Cref{fig:asd_composite}C, left). For the on-site block itself, although
$K_{ii}$ is parametrised as a full $3\times3$ matrix, the learned operator is
approximately diagonal in the spin components, with in-plane entries
$K_{xx}\approx K_{yy}\equiv K_\parallel$, out-of-plane entry $K_{zz}\equiv K_\perp$, and negligible off-diagonal coupling. Both coupling constants depend approximately evenly on $m_z$ and peak near the disordered state ($m_z\approx0$), reflecting the reduced phase space available for fluctuations in the ordered phases (\Cref{fig:asd_composite}C, middle).
Finally, the recovered spatial interaction kernel exhibits a dominant nearest-neighbour structure consistent with the lattice symmetry, recovering the short-range exchange coupling without prescribing this structure (\Cref{fig:asd_composite}C, right).
The recovered thermodynamic components provide an interpretable description of the observed relaxation.

\paragraph*{Layered
CrBr\textsubscript{3} from a learned fault interaction to multistep field response}
CrBr\textsubscript{3} is a van der Waals ferromagnet whose interlayer magnetic coupling depends on the stacking registry of adjacent layers. Different stacking configurations, including monoclinic M-stacking and rhombohedral AB-stacking, correspond to different relative translations between adjacent layers and can favour antiferromagnetic or ferromagnetic interlayer correlations (\Cref{fig:stacking_fault}A)~\cite{yang_controlling_2023,yao_multiple_2023}.
Experimentally, such magnetic switching can be probed through magneto-photoluminescence (magneto-PL), where photoluminescence refers to light emitted following optical excitation and its response can depend on the magnetic state of the material~\cite{Grzeszczyk2023ExcitonMagnetization}. Our magneto-PL measurements on an approximately $50\,\mathrm{nm}$ CrBr$_3$ flake show abrupt changes in PL intensity associated with magnetic switching (\Cref{fig:stacking_fault}D). These measurements reveal the occurrence of magnetic switching, but do not directly resolve the spatial magnetisation dynamics that connect the local stacking fault to the observed macroscopic response. In particular, how the modified exchange interface reorganises magnetic domains and mediates switching across the film remains difficult to access experimentally.
Atomistic spin dynamics simulations can in principle resolve these processes, but become prohibitively expensive over the spatial and temporal scales required to follow domain formation and collective switching.
This motivates the use of \method{} to learn an effective mesoscopic description that retains the influence of the microscopic stacking-dependent exchange while enabling direct study of the resulting domain evolution and macroscopic magnetisation dynamics.

Because the layered geometry provides prior information about the distinction between in-plane and interlayer interactions, we impose additional structure on the interaction kernel, separating these two contributions to reduce the parameter space and make their physical roles explicit. We generate atomistic spin dynamics trajectories for a $51.4\,\mathrm{nm}$ CrBr$_3$ film with a single antiferromagnetic stacking fault at its midplane. Averaging three atomic layers per slab gives 28 resolved slabs. From
these trajectories, \method{} learns the energy-density function, the spatial
interaction, and the dissipation operator of the coarse-field equation.
We represent the spatial interaction kernel as a separable product: its layer-dependent factor captures variations associated with surfaces and stacking faults, whereas its in-plane factor represents short-range exchange. At coarse indices
$(x,y,z)$, the interaction kernel is
\[
J_{(x,y,z),(x',y',z')}
=J_z(z,z')\,J_{xy}(x-x',y-y')
\equiv
(J_z\otimes J_{xy})_{(x,y,z),(x',y',z')}.
\label{eq:crbr3_tensor_kernel}
\]
Here $J_z$ is a global layer-to-layer matrix, whereas $J_{xy}$ is a local
finite-range in-plane kernel.

Applying \method{} to these trajectories yields the interaction kernels and free-energy landscape shown in \Cref{fig:stacking_fault}B and C. The learned interlayer kernel $J_z$ has a predominantly tridiagonal structure, with the coupling across the stacking fault between coarse-grained layers 13 and 14 reversing sign relative to the bulk interlayer coupling (\Cref{fig:stacking_fault}B, left). The in-plane kernel $J_{xy}$ is isotropic and approximately nearest-neighbour, consistent with the symmetry of the CrBr$_3$ lattice (\Cref{fig:stacking_fault}B, right). Through the learned free energy, this fault-induced change in the interaction kernel alters the equilibrium structure of the system. Restricting the free energy to configurations with spatially uniform magnetisation within the blocks above and below the fault reveals four distinct minima in the mean-magnetisation and block-contrast coordinates (\Cref{fig:stacking_fault}C). Two correspond to parallel block magnetisations and two to antiparallel block magnetisations. This contrasts with a uniformly ferromagnetic system, which exhibits only two symmetry-related minima. The additional basins therefore reveal fault-induced equilibrium states and provide a mesoscopic free-energy picture for the multiple magnetic transitions associated with the stacking fault.

\clearpage
\refstepcounter{figure}
\label{fig:stacking_fault}
\begin{center}
  \includegraphics[width=\textwidth]{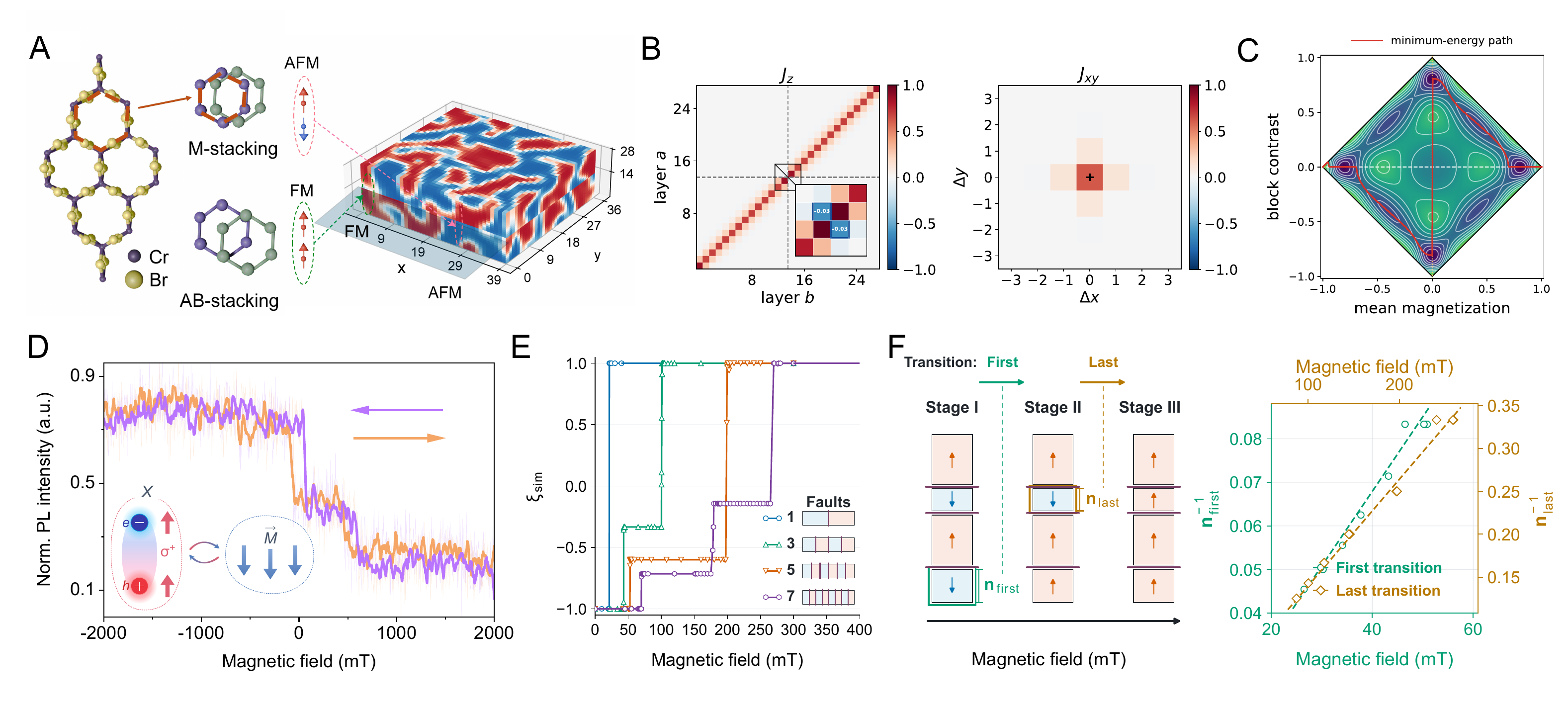}
\end{center}
\noindent\textbf{Figure~\thefigure.} \textbf{A learned stacking-fault interaction connects internal geometry to multistep magnetic response in CrBr\textsubscript{3}.}
    \textbf{A.} Structural and magnetic schematic of layered CrBr\textsubscript{3}.
    Left: crystal structure of CrBr\textsubscript{3} (Cr atoms in purple, Br atoms in gold).
    Middle: top and side views of representative M-stacking and AB-stacking configurations, where M denotes monoclinic stacking and AB denotes rhombohedral stacking and the two adjacent layers are coloured purple and green.
    Right: schematic AFM (antiparallel) and FM (parallel) interlayer spin alignments across a stacking fault, shown alongside the simulated three-dimensional magnetisation field $m_z$ at the final time step.
    The light-blue plane marks the stacking-fault interface between coarse slabs 13 and 14 (zero-based indices), and the pink (AFM) and green (FM) dashed circles highlight representative anti-correlated and correlated magnetic subdomains on either side of the fault.
    \textbf{B.} Effective interaction kernels learned by \method{} without supplying the fault location as a model input.
    Left: learned interlayer kernel $J_z$ as a function of layer indices $a$ and $b$, showing a predominantly tridiagonal structure. The coupling across the stacking-fault interface between coarse slabs 13 and 14 (zero-based indices) has the opposite sign to the bulk interlayer coupling.
    Right: learned in-plane kernel $J_{xy}$ as a function of in-plane displacement $(\Delta x,\Delta y)$, showing an isotropic, short-range profile.
\textbf{C.} Learned free-energy landscape for configurations with spatially uniform magnetisation within each of the two blocks. The coordinates are the mean magnetisation $m=(m_{\mathrm{top}}+m_{\mathrm{bottom}})/2$ and block contrast $\Delta m=(m_{\mathrm{top}}-m_{\mathrm{bottom}})/2$.
    The four minima represent correlated states at $(m,\Delta m)=(\pm m_0,0)$ and anti-correlated states at $(m,\Delta m)=(0,\pm m_0)$.
    The red curves indicate minimum-energy paths connecting these basins.
\textbf{D.} Magneto-photoluminescence signature of magnetic transitions on a roughly $50\,\mathrm{nm}$ CrBr$_3$ flake. The normalised PL intensity is shown as a function of out-of-plane magnetic field for sweeps from $-2\,\mathrm{T}$ to $2\,\mathrm{T}$ and from $2\,\mathrm{T}$ to $-2\,\mathrm{T}$, with a step-like PL response to the applied magnetic field. The purple and orange curves show smoothed traces, while the faint traces in the background show the corresponding unsmoothed experimental data. Arrows indicate the magnetic-field sweep directions. The inset shows a schematic of exciton--magnetisation coupling in CrBr$_3$. $X$ denotes a CrBr$_3$ exciton, represented here by an electron--hole pair with both carrier spins oriented upward. The $\sigma^{+}$ label denotes the circular polarisation associated with this representative optically active exciton configuration. The local magnetisation vector $\mathbf{M}$ is chosen to point downward.
    \textbf{E.} Calculated equilibrium mean interfacial sign correlation $\xi_{\mathrm{sim}}$ versus external field for the $51.4\,\mathrm{nm}$ film containing one, three, five, or seven approximately uniformly spaced stacking faults. This observable averages the product of longitudinal magnetisation signs across each fault over in-plane cells, fault planes, and equilibrium configurations [\eqref{eq:sign_corr_sim}]. The plateaus correspond to dominant block-alignment sectors. The small diagrams show the fault locations.
\textbf{F.} Dependence of the first and final transition fields on fault positions for films containing three faults. Left: starting from alternating block magnetisations, the outer block opposing the field reverses first, followed by the remaining antiparallel internal block. Green and orange outlines identify these blocks, whose thicknesses are $n_{\mathrm{first}}$ and $n_{\mathrm{last}}$, respectively, measured in resolved slabs. Right: inverse thicknesses $n_{\mathrm{first}}^{-1}$ and $n_{\mathrm{last}}^{-1}$ plotted against the first and final transition fields, respectively. Symbols denote the simulated geometries; dashed linear fits show the approximately linear relationships.
\par
\vspace{1.5\baselineskip}

Having identified how a single stacking fault reshapes the free-energy landscape and introduces additional equilibrium basins, we next examine how this mechanism extends to films containing multiple faults. In particular, we ask how the number and spatial arrangement of stacking faults modify the field-dependent magnetic response. Addressing this question requires systematic variation of the fault configuration while resolving the resulting magnetisation dynamics. Atomistic spin dynamics simulations can represent individual faults explicitly, but systematic exploration over fault number, position, film geometry, and applied field becomes prohibitively expensive as the number of microscopic spins increases. Experiments can measure the resulting macroscopic switching response, but the number and positions of stacking faults are difficult to control independently, and the internal magnetisation dynamics associated with a particular fault configuration are not directly resolved. Once \method{} has learned the mesoscopic free energy and interaction structure, however, different fault configurations can be introduced through the layer-dependent coupling and explored directly at the coarse-grained level.

We introduce an out-of-plane field through the Zeeman term
\[
\mathcal{F}_{\theta,J,h}[\mathbf m]
=
\mathcal{F}_{\theta,J}[\mathbf m]
-
\alpha_h h \int_\Omega m_z(\mathbf x)\,d\mathbf x,
\]
where $h$ is the dimensionless applied field, $m_z$ is the out-of-plane magnetisation component, and $\alpha_h$ converts the field to the learned free-energy scale. We calibrate $\alpha_h$ after training as described in Materials and Methods. To vary the fault geometry, we retain the 28 resolved slabs and insert translated copies of the learned fault neighbourhood at prescribed interfaces of $J_z$, keeping the other learned quantities fixed. We quantify the response by the mean sign correlation across all fault planes, so that each internal interface contributes to the observable:
\begin{equation}
\xi_{\mathrm{sim}}(h)
=
\mathbb E_h\!\left[
\frac{1}{N_fN_{xy}}
\sum_{f=1}^{N_f}
\sum_{i=1}^{N_{xy}}
\operatorname{sign}\!\left[m_{z,i}^{f,+}\right]
\operatorname{sign}\!\left[m_{z,i}^{f,-}\right]
\right],
\label{eq:sign_corr_sim}
\end{equation}
where $N_f$ is the number of stacking faults, $N_{xy}$ is the number of
in-plane coarse-grained cells, and $m_{z,i}^{f,+}$ and $m_{z,i}^{f,-}$ denote
the out-of-plane magnetisation in cell $i$ of the slabs immediately above and
below fault $f$ in a configuration at applied field $h$. The expectation $\mathbb E_h$ averages over equilibrium configurations within and across the retained block-alignment sectors at fixed $h$. Thus, $\xi_{\mathrm{sim}}=1$ and $-1$ indicate equal and opposite longitudinal magnetisation signs, respectively, at every interface cell throughout the equilibrium ensemble. Intermediate values can arise from mixed alignments within a fault plane, different alignments among fault planes, or averaging over configurations. Direct trajectories do not sample all relevant block-alignment sectors on the accessible time scale. We therefore use bidirectional annealed importance sampling, which connects ensembles through intermediate distributions and estimates their relative equilibrium weights from bidirectional switching work~\cite{neal2001annealed,shirts2003equilibrium}, to evaluate \eqref{eq:sign_corr_sim}. Computational details are given in Materials and Methods and the SI Appendix, section~4D.
\par
We vary the interlayer interaction kernel $J_z$ to represent one, three, five, and seven approximately uniformly spaced faults while keeping the other learned quantities fixed (SI Appendix, section~4D). The resulting equilibrium field responses exhibit distinct plateaus that reflect successive changes in the dominant block-alignment sectors (\Cref{fig:stacking_fault}E). Increasing the fault number changes the plateau sequence and extends the response to higher fields in the tested geometries. The simulated plateaus reflect changes in interfacial block alignment, whereas the PL steps in \Cref{fig:stacking_fault}D are an optical signature of changes in magnetic state. These calculations provide a possible magnetic interpretation of our PL steps.

We next fix the number of stacking faults at three and examine how their positions affect the first and final equilibrium transition fields (\Cref{fig:stacking_fault}F). Each fault geometry divides the 28-slab film into four magnetic blocks with thicknesses $n_1,n_2,n_3,n_4$, counted consecutively through the film in resolved slabs, with $\sum_{k=1}^{4}n_k=28$.  The left schematic illustrates two successive transitions from alternating block magnetisations as the field increases. The outer block opposing the field reverses first; its thickness is $n_{\mathrm{first}}$, and it is outlined in green. The remaining antiparallel internal block, outlined in orange, then reverses to align all four blocks with the field. Its thickness is $n_{\mathrm{last}}=\min(n_2,n_3)$. The outer block that reverses first is separated by one block from the internal block that reverses last.

Across the tested fault geometries, the first and final transition fields vary approximately linearly with $n_{\mathrm{first}}^{-1}$ and $n_{\mathrm{last}}^{-1}$, respectively, as shown in the right panel. The scaling can be understood from the competition between the Zeeman energy gained by aligning a block with the applied field and the interfacial free-energy cost of replacing antiparallel alignments with parallel ones. For a fixed lateral area and comparable magnetisation magnitude, the Zeeman energy gain scales with the applied field times the thickness of the reversing block. If the interfacial cost remains comparable across fault geometries, the critical field therefore scales approximately inversely with the block thickness, so that thicker blocks switch at lower applied fields.

%% file: sections/05_discussion.tex
\section*{Discussion}
\subsection*{Physical interpretation and computational scale}
The results demonstrate three uses of learned mesoscopic dynamics: extending simulations in time and space, analysing thermodynamic structure, and isolating how spatial interactions control collective magnetic response.

First, evolving the coarse field avoids resolving every atomistic spin and extends short microscopic observations to longer-time simulations. In the two-dimensional atomistic spin system, predictions agree with an atomistic reference up to $1\,\mathrm{ns}$ and remain bounded during relaxation to $25\,\mathrm{ns}$. For the two-dimensional ferromagnet and CrBr\textsubscript{3}, the wall-clock speedups over atomistic spin dynamics are approximately $4.1\times10^{3}$ and $4.0\times10^{4}$, respectively, without operator-specific optimisation (Materials and Methods). The Ising calculations also demonstrate evaluation on larger spatial domains at fixed coarse-graining scale and resolution (SI Appendix, section~2G).

Second, the learned equations preserve thermodynamic structure while allowing the free energy, interaction kernel, and dissipation to be analysed separately. In Ising, the learned energy density reproduces the double-well and single-well profiles at the two temperatures, and the conditional dissipation response agrees with the finite-block fluctuation reference. In the atomistic systems, the energy density reveals uniaxial anisotropy, the dissipation varies with the magnetisation, and the CrBr$_3$ interaction kernel identifies a reversal of interlayer coupling at the stacking fault.

Third, the learned interactions enable controlled calculations of mesoscopic magnetic properties. In CrBr$_3$, changing the number and placement of fault interfaces while holding the other learned quantities fixed isolates their effects on interfacial alignment and field-induced switching. The calculated plateaus track changes in magnetic alignment across fault interfaces, providing a magnetic perspective on the multistep optical response observed in magneto-photoluminescence. In the future, atomic-resolution scanning transmission electron microscopy~\cite{yang_controlling_2023} combined with magnetic measurements could help examine the predicted relation between stacking-fault geometry and magnetic response in real samples.

\subsection*{Relation to other data-driven methods}
We then discuss related work on data-driven dynamics modelling.
Neural operators such as DeepONet and Fourier neural operators learn black-box mappings between input and solution fields~\cite{lu2021learning,li2020fourier}. Equation-discovery methods including SINDy, AI-Feynman, and PDE-Net infer governing expressions using prescribed candidate libraries, symbolic search spaces, or structured differential representations~\cite{brunton2016discovering,udrescu2020ai,long2018pde,long2019pde}. For coarse-grained particle trajectories, symmetry-informed candidate libraries and sparse regression have been used to infer deterministic hydrodynamic equations for active matter~\cite{supekar2023learning}. Effective stochastic dynamics can also be learned from microscopic simulations by fitting drift and diffusion functions using objectives based on stochastic numerical integrators~\cite{dietrich2023learning}. OnsagerNet, generalised-Onsager learning, and GENERIC-informed models parametrise thermodynamically structured dynamics for finite-dimensional state vectors rather than directly parametrising mesoscopic field equations~\cite{yu2021onsagernet,chen2024constructing,zhu2025identifiable,grmela1997dynamics,zhang2022gfinns}. Recent work extends Onsager-structured learning to mesoscopic field dynamics through deterministic trajectory fitting~\cite{li2026hypothesisdrivenconstructionmesoscopicdynamics}.
A related approach learns conserved density dynamics from molecular dynamics trajectories~\cite{chen2026learning}. By contrast, DeepMesoD develops a more general mesoscopic modelling framework based on symmetry principles and the fluctuation--dissipation relation, with flexible free-energy functionals and dissipation operators for nonconserved fields.

Stat-PINNs and DeepMesoD both use fluctuation statistics to constrain dissipation and mean evolution to learn the free energy~\cite{huang2025statpinn}, but obtain these statistics differently. Stat-PINNs prepare and evolve microscopic replica ensembles around prescribed field profiles, whereas DeepMesoD uses increments along recorded mesoscopic field trajectories. The same trajectories supply both the mean evolution and fluctuation statistics, removing the need to prepare and evolve additional microscopic replicas for each prescribed field profile. The two methods also differ in how they represent operators. Stat-PINNs learn weak-form dissipation entries in a selected finite-element space. DeepMesoD defines a factor $\mathcal B$ on the field space and constructs $\mathcal K=\mathcal B\mathcal B^*$, then learns its numerical representation. This separates the definition of the operator structure from its numerical discretisation. Self-adjointness and positive semidefiniteness follow from the factorisation. DeepMesoD also learns the spatial interaction kernel $J(\mathbf x,\mathbf y)$, which reveals how different regions interact. In CrBr$_3$, modifying this kernel enables controlled numerical experiments that help reveal how spatial interactions shape collective magnetic behaviour (\Cref{fig:stacking_fault}E,F). DeepMesoD includes the thermal drift $\operatorname{Div}_{\phi}\mathcal K[\phi]/(\beta V_c)$ in the It\^o equation~\eqref{eq:spde}, whereas Stat-PINNs derive this term but omit it under their asymptotic scaling. In the formal Fokker--Planck equation, this term ensures zero probability current at Gibbs equilibrium. At finite coarse-graining volume $V_c$, omitting it while keeping the same free energy and noise covariance generally breaks detailed balance with respect to that equilibrium.

\subsection*{Dynamical assumptions and scope}
Hohenberg and Halperin classify critical dynamics according to whether the order parameter is conserved and how it couples to other slow variables~\cite{HohenbergHalperin1977}. Among these classes, Model A describes relaxation of a nonconserved order parameter, whereas Model B describes transport-driven evolution of a conserved order parameter. Couplings to additional slow fields and reversible dynamics give rise to further classes. Here, DeepMesoD describes the selected coarse field using Markovian Model-A-type dynamics at the prescribed scale.

Atomistic LLG dynamics includes reversible precession as well as dissipation. At the chosen coarse-graining scale, spatial averaging suppresses incoherent transverse precession, and the SI diagnostics~(sections~3B and~4A) show no statistically resolved reversible coupling between the remaining transverse motion and the longitudinal magnetisation $m_z$. These observations support the dissipative description of the longitudinal dynamics compared here. Conserved fields or systems with appreciable reversible couplings require a different field equation. Eliminating unresolved microscopic variables can produce memory and coloured noise, as described by Mori--Zwanzig theory~\cite{Mori1965,zwanzig2001nonequilibrium}. The ab initio generalised Langevin equation provides a data-driven example in which a force field, memory kernel, and noise model are learned from atomistic trajectories while satisfying the fluctuation--dissipation relation~\cite{xie2024aigle}. These are worthy avenues to explore in future work.

Beyond these dynamical assumptions, the accuracy of the learned mesoscopic equation also depends on the microscopic trajectories used as reference data. The trajectories studied here are generated from effective spin Hamiltonians rather than first-principles simulations.
Machine-learned interatomic potentials could extend this framework to more materials by providing microscopic trajectories with first-principles-informed energies and forces~\cite{friederich2021machine,jia2020pushing,guo2022extending}. Spin-dependent potentials such as DeepSPIN also provide magnetic torques for spin-lattice dynamics~\cite{yang2024deepspin}. Learning mesoscopic equations from these trajectories could extend simulations of collective magnetic dynamics towards device-relevant spatial scales.

%% file: sections/06_materials_and_methods.tex
\matmethods{

\subsection*{Microscopic data generation}
Ising--Kac trajectories are generated by heat-bath Glauber dynamics on a $4{,}096\times4{,}096$ periodic lattice with spacing $1/4{,}096$. The Gaussian Kac interaction has range $\epsilon_{\mathrm K}=1/64$, corresponding to 64 lattice spacings. The reported datasets use zero external field at $T=0.75$ and $T=1.5$, with multiscale Gaussian-random-field initial conditions.

We generate atomistic spin dynamics trajectories for the Cr\textsubscript{2}Ge\textsubscript{2}Te\textsubscript{6}-like two-dimensional ferromagnet by integrating the stochastic LLG equation with VAMPIRE v7.0.0~\cite{evans2014atomistic}. Ten independent trajectories of a $1{,}024\times1{,}024$ monolayer are generated at $T=10$\,K for $250$\,ps, with configurations saved every $1$\,ps.

The same VAMPIRE package is used to generate atomistic spin dynamics trajectories for a $51.4$\,nm CrBr$_3$ film with one antiferromagnetic stacking fault at its midplane. The $200\times200$\,nm$^2$ reference film contains approximately $2\times10^7$ spins. Ten independent zero-field trajectories are generated at $T=50$\,K for $500$\,ps, with full spin configurations saved every $2$\,ps. The SI Appendix~(sections~2A, 3A, and~4A) gives the microscopic parameters, integration settings, and initial conditions for all three systems.

\subsection*{Common coarse-graining and learning procedure}
The    observations $z_i(t)$ at positions $\mathbf x_i$  are block averages over    neighbourhoods  containing $V_c$ microscopic sites. For Ising--Kac, a block width of 32 gives $V_c=1{,}024$ and maps the $4{,}096^2$ lattice to a $128^2$ array of coarse    magnetisations . For the two-dimensional ferromagnet, a block width of 16 gives $V_c=256$ and maps the $1{,}024^2$ lattice to a $64^2$ array. For CrBr\textsubscript{3}, the atomistic spin configurations are averaged over nonoverlapping blocks of size $5.01\times5.42\times1.84$\,nm$^3$, each containing three atomic layers. This procedure produces a $40\times37\times28$ array, with $V_c=480$.

The reported calculations place the coarse observations on regular arrays and use a unit-spacing piecewise-constant numerical representation. These control regions define the spatial quadrature. They are distinct from the microscopic averaging    neighbourhoods  that determine $V_c$.

The observations satisfy $z_i(t)\approx\phi(\mathbf x_i,t)$. In the training formulas below, $z(t)$ denotes these observations arranged by position and field component. The subscript $N$ on numerical quantities denotes discretisation on the $N$ coarse cells. The energy-density function $\rho$ is represented by a multilayer perceptron, written as $\rho_\theta$ when its trainable parameters $\theta$ are explicit. The interaction kernel $J$ is represented by trainable coefficients $J_N$, which act identically on each field component to form the interaction input. Applying the spatial quadrature to \eqref{eq:learned_free_energy} gives the numerical free energy $\mathcal F_N(z)$. A convolutional neural network with parameters $\vartheta$ represents the local dependence of the factor kernel $b$: it takes the field values in a    neighbourhood  of each cell and returns a $q\times p$ coefficient matrix for each allowed noise-source offset. Assembling these coefficients gives $B_{\vartheta,N}(z)\in\mathbb R^{Nq\times Np}$ and the dissipation operator $K_{\vartheta,N}(z)=B_{\vartheta,N}(z)B_{\vartheta,N}(z)^{\mathsf T}$.

Training uses three stages. Stage 1 fits an auxiliary conditional drift $\widehat f$ by    minimising  the mean-squared error $\mathbb E_{\mathrm{train}}\lVert\widehat f(z^n)-d_N^n\rVert^2$, where $d_N^n$ is a finite-difference drift label specified for each system in the SI Appendix, sections~2A and~2D, 3C.1, and~4B.2. At consecutive recorded times $t_n$ and $t_{n+1}$, write $\Delta t_n=t_{n+1}-t_n$ and $z^n=z(t_n),\quad\Delta z^n=z^{n+1}-z^n$. Stage 2 fits the factor parameters using the residual increments $r_N^n=\Delta z^n-\widehat f(z^n)\Delta t_n$. Let $W$ index the magnetisation components in a group of neighbouring coarse cells. Stage 2    minimises  the Gaussian quasi-likelihood
\[
\begin{aligned}
\mathcal L_K
&=\frac{1}{2|W|}\Big\langle
\log\det(\Sigma_W^n+\lambda I_W)
+(r_{N,W}^n)^{\mathsf T}(\Sigma_W^n+\lambda I_W)^{-1}r_{N,W}^n
\Big\rangle_{n,W},\\
\Sigma_W^n
&=\frac{2\Delta t_n}{\beta V_c}[K_{\vartheta,N}(z^n)]_{W,W}.
\end{aligned}
\]
Here $\Sigma_W^n$ is the model covariance of the residual increments indexed by $W$, $r_{N,W}^n$ is the residual restricted to $W$, $|W|$ is the number of scalar field entries indexed by $W$, $I_W$ is its identity matrix, $\lambda>0$ is a numerical covariance floor, and the brackets average over sampled time-window pairs.

After Stage 2, the fitted factor parameters are held fixed. The resulting operator $K_{\vartheta,N}(z)$ remains state dependent while Stage 3 jointly    optimises  the free energy and interaction against the observed field changes. Its direct drift loss is
\[
\begin{aligned}
e_{F,J,i}^n
&=\left[\frac{\Delta z^n}{\Delta t_n}\right]_i
+\left[K_{\vartheta,N}(z^n)\nabla_{z}\mathcal F_N(z^n)\right]_i
-\frac{1}{\beta V_c}
\left[\operatorname{Div}_{z}K_{\vartheta,N}(z^n)\right]_i,\\
\mathcal L_{F,J}^{\mathrm{dir}}
&=\mathbb E_{\mathrm{train}}\!\left[\left|e_{F,J,i}^n\right|^2\right].
\end{aligned}
\]
Here $i$ indexes coarse-cell magnetisation components, and $\mathbb E_{\mathrm{train}}$ denotes the empirical average over sampled entries and increments. The coordinate state divergence and the system-specific sampling and loss    normalisations  are given in the SI Appendix, sections~1C, 2D, 3C, and~4B.
\subsection*{Ising--Kac implementation}
The Ising free energy uses a homogeneous density $g_\theta$, the parameterised version of $g$ in the Results, and a symmetric, unit-sum interaction $\widehat J_N$ with offsets in a $15\times15$ neighbourhood. The scalar factor reads a $15\times15$ field patch and acts on noise sources at the nine offsets in a $3\times3$ neighbourhood. Its coefficients are even under global spin reversal.

\subsection*{Two-dimensional-ferromagnet implementation}
The two-dimensional ferromagnet uses a component-shared $13\times13$ unit-sum interaction, a three-component drift network, and a scalar energy-density network. The factor has $q=p=3$ and assigns a $3\times3$ matrix to each offset in $\mathcal S_B=\{-1,0,1\}^2$. Three circular convolutional layers give a $7\times7$ state receptive field, while the noise sources lie in a $3\times3$ neighbourhood.

\subsection*{CrBr\textsubscript{3} implementation and equilibrium calculations}
The CrBr$_3$ drift and free energy use independently fitted separable interactions: a unit-sum $7\times7$ in-plane kernel and a symmetric, unnormalised $28\times28$ layer matrix. The layer matrix carries the interlayer coupling strength. The factor acts within each slab and shares its coefficients across slabs; interlayer coupling enters through the free energy. Stage 3 uses the unnormalised direct drift loss. System-specific free-energy formulas, network architectures, sampling, and optimisation settings are given in the SI Appendix, section~4B. Because the training trajectories are generated at zero field, we calibrate the coupling to the applied field after training. We add the Zeeman density $-\alpha_h h m_z$ and fit $\alpha_h$ from paired opposite-field increments at $\pm25$ and $\pm50$~mT. The fit gives $\alpha_h=0.1894$ for $h=B/(\mu_0M_s)$ with $\mu_0M_s=0.3393$~T, where $M_s$ is the saturation magnetisation. The learned energy density, interaction kernels, and dissipation operator remain fixed during this calibration.

Multi-fault calculations retain the 28-slab film thickness. We decompose the learned single-fault matrix into a background and a local fault correction, then add translated copies of this correction at the prescribed interfaces. Overlapping contributions are summed, and entries outside the film are omitted. The construction and example matrices are given in the SI Appendix, section~4D. The energy-density network, in-plane kernel, dissipation factor, temperature, and field calibration remain fixed.

Long direct trajectories do not cross all block-alignment sectors on the available simulation time scale. Equilibrium sector weights are therefore computed by bidirectional annealed importance sampling. For each sampled configuration, we evaluate the sign statistic inside \eqref{eq:sign_corr_sim} and combine its sector averages using the restricted partition-function weights. A transition bracket is formed by two adjacent simulated field values for which the dominant block-alignment sector changes. The first and last brackets locate the two transitions plotted in \Cref{fig:stacking_fault}F. Further sampling details are given in the SI Appendix, section~4D.

\subsection*{Simulation timing}
For the Cr\textsubscript{2}Ge\textsubscript{2}Te\textsubscript{6}-like two-dimensional ferromagnet, advancing the system by 1 ps takes 16.2 s with VAMPIRE~\cite{Skubic2008,evans2014atomistic} on 128 CPU cores, and 3.96 ms with \method{} on one NVIDIA L40S GPU. For CrBr\textsubscript{3}, advancing the system by 2 ps takes 190.05 s with VAMPIRE~\cite{Skubic2008,evans2014atomistic} on 128 CPU cores, and 4.8 ms with \method{} on the same GPU.

\subsection*{Magneto-photoluminescence (PL) measurements}
Magneto-PL measurements were performed in a home-built backscattering microscope integrated with an attocube attoDRY 2100 closed-cycle cryostat operated at $1.7\,\mathrm{K}$ and equipped with a superconducting magnet providing fields up to $9\,\mathrm{T}$. The sample was excited with a continuous-wave $730\,\mathrm{nm}$ laser at a power of $\approx 50\,\mu\mathrm{W}$ and focused to a spot size of $\approx 1\,\mu\mathrm{m}$, and the emitted light was collected through the same objective and detected using a spectrometer equipped with a 150 lines/mm grating and a CCD camera. An out-of-plane magnetic field was applied and swept in both directions. Circularly polarised excitation was generated using a quarter-wave plate placed in the excitation path.

\subsection*{Data and code availability}
The code is available at \href{https://github.com/MLDS-NUS/DeepMesoD}{https://github.com/MLDS-NUS/DeepMesoD}.
}

%% file: SI/sections/01_theory_and_methodology.tex
\section{{Theoretical and methodological details}}

\subsection{{Function-space formulation}}

\subsubsection{Free-energy derivative}

We use the field space \(\Phi=L^2(\Omega;\mathbb R^q)\) and the interaction operator \(\mathcal L_J\) defined in the main text. The inner product is $\langle u,v\rangle_\Phi=\int_\Omega u(\mathbf x)\cdot v(\mathbf x)\,d\mathbf x$. The dual of \(\Phi\) is identified with \(\Phi\) through the \(L^2\) Riesz representation. The functional derivative therefore satisfies $D\mathcal F[\phi](v)=\langle\delta\mathcal F[\phi]/\delta\phi,v\rangle_\Phi$ for every perturbation $v\in\Phi$.

Let \(\rho_\theta\colon\mathbb R^q\times\mathbb R^q\to\mathbb R\) be the continuously differentiable energy-density function, with \(\theta\) denoting its trainable parameters. For \(\psi=\mathcal L_J\phi\), write
\[
\mathcal F_{\theta,J}[\phi]
=\int_\Omega\rho_\theta\!\left(\phi(\mathbf x),\psi(\mathbf x)\right)\,d \mathbf x.
\]
We assume that this functional is Fr\'echet differentiable on the fields considered. With \(u_1(\mathbf x)=\partial_1\rho_\theta(\phi(\mathbf x),\psi(\mathbf x))\) and \(u_2(\mathbf x)=\partial_2\rho_\theta(\phi(\mathbf x),\psi(\mathbf x))\), its derivative is
\[
{\begin{aligned}
D\mathcal F_{\theta,J}[\phi](v)
&=\langle u_1,v\rangle_\Phi
+\langle u_2,\mathcal L_Jv\rangle_\Phi,\\
\frac{\delta\mathcal F_{\theta,J}[\phi]}{\delta\phi}
&=u_1+\mathcal L_J^*u_2.
\end{aligned}}
\]
Here \(\mathcal L_J^*\) is the Hilbert-space adjoint. Under the unit-integral convention in the main text,
\[
(\mathcal L_J^*u)(\mathbf x)=u(\mathbf x)-\int_\Omega J(\mathbf y,\mathbf x)u(\mathbf y)\,d \mathbf y.
\]

\subsection{{Finite-scale stochastic equation}}

The    white noise $\xi$  in the    main text is the formal time derivative of  a cylindrical Wiener process  $W_t$  on $\mathcal U=L^2(\Omega;\mathbb R^p)$.  Its components satisfy $\mathbb E[\xi_a(\mathbf x,t)\xi_c(\mathbf y,s)]=\delta_{ac}\delta(\mathbf x-\mathbf y)\delta(t-s)$. Thus the term $\mathcal B[\phi]\xi$ is interpreted through the It\^o integral $\int\mathcal B[\phi(t)]\,dW_t$.  For each admissible field $\phi$, the factor kernel satisfies $b[\phi]\in L^2(\Omega\times\Omega;\mathbb R^{q\times p})$, so its integral operator $\mathcal B[\phi]\colon\mathcal U\to\Phi$ is Hilbert--Schmidt.

For the field equation in the main text, let \(\{e_n\}_{n\geq1}\) be an orthonormal basis of \(\Phi\). If \(\mathcal K\) is Fr\'echet differentiable and the following limit converges in \(\Phi\) independently of the chosen basis, we define
\[
\operatorname{Div}_{\phi}\mathcal K[\phi]
=\lim_{M\to\infty}
\sum_{n=1}^{M}
D\mathcal K[\phi][e_n]e_n.
\]
Here \(D\mathcal K[\phi][e_n]\) is the derivative in the direction \(e_n\). Along an adapted field trajectory, we assume that \(\mathcal B[\phi(t)]\) is predictable and that \(\mathbb E\int_0^{t_*}\|\mathcal B[\phi(t)]\|_{\mathrm{HS}}^2\,dt<\infty\) for every finite \(t_*>0\). These conditions make the stochastic integral well defined.

\subsection{{Finite-coordinate stationary density}}

Here $N$ is the number of coarse cells and $q$ the number of field components per cell.

The piecewise-constant reconstruction $\mathcal R_N\colon\mathbb R^{Nq}\to\Phi$ assigns each cell its field value and defines the numerical field space $\Phi_N=\operatorname{im}\mathcal R_N$. For the coordinate calculations below, $u\in\mathbb R^{Nq}$ collects the field values by position and component. When evaluating training losses, these coordinates are supplied by the observations $z$ defined in Materials and Methods. Their piecewise-constant reconstruction is $\phi_N=\mathcal R_Nu$. The numerical free energy $\mathcal F_N$ evaluates the free-energy functional defined in the main text with unit spatial integration weights. These weights give identity mass matrices, so the coordinate gradient represents the $L^2$ free-energy derivative. In another numerical basis, the mass matrices must be included when representing this derivative, the factor adjoint, and the noise covariance.

 For numerical efficiency, we represent $J(\mathbf x,\mathbf y)$ as piecewise constant on pairs of numerical cells and learn its matrix of coefficients $J_N$. These coefficients can be shared across spatial offsets where translation symmetry applies. This is a numerical representation of the interaction kernel; a general function of $(\mathbf x,\mathbf y)$ can also be parameterised when needed.

 The factor coefficients are numerical coefficients in the state and noise bases. Let $\mathcal S_B$ be the allowed noise-source offsets, and let    $P_i^Ru$  denote the field patch of radius    $R$ centred  at cell $i$ , where $R$ bounds both the state dependence and the noise-source range as in the main text . For a latent vector    $\eta_N\in\mathbb R^{Np}$ , the factor acts as
\[
[B_{\vartheta,N}(u)   \eta _N]_{i,a}
=\sum_{s\in\mathcal S_B}\sum_{c=1}^{p}
b_{\vartheta,ac}(P_i   ^R u,s)(   \eta _N)_{i+s,c}.
\]
Here $a\in\{1,\ldots,q\}$ indexes field components and $c\in\{1,\ldots,p\}$ indexes noise channels. On a periodic homogeneous domain, the same coefficient network is applied at every position. With fitted parameters, write $B_N(u)=B_{\vartheta,N}(u)$ and $K_N(u)=B_N(u)B_N(u)^{\mathsf T}$. The operator acts on a vector $g$ as $K_N(u)g=B_N(u)(B_N(u)^{\mathsf T}g)$ without forming a dense covariance matrix. Since two field entries can share a noise source, the support of $K_N$ lies in $\mathcal S_B-\mathcal S_B$ and    is bounded by radius $2R$ .

The corresponding finite-coordinate equation is
\begin{equation}
du
=\left[-K_N(u)\nabla_u\mathcal F_N(u)
+\frac{1}{\beta V_c}\operatorname{Div}_uK_N(u)\right]dt
+\sqrt{\frac{2}{\beta V_c}}B_N(u)\,dW_t^{(N)}.
\label{eq:si_coordinate_ito_equation}
\end{equation}
where $W_t^{(N)}$ is a standard $Np$-dimensional Wiener process and
\[
[\operatorname{Div}_uK_N(u)]_i=\sum_{j=1}^{Nq}\partial[K_N(u)]_{ij}/\partial u_j.
\]

Assume that the finite-coordinate equation is nonexplosive and admits a density satisfying the Fokker--Planck equation. For the complete drift, its probability current reduces to
\[
j_i
=-\sum_j[K_N]_{ij}\left[(\partial_{u_j}\mathcal F_N)\pi
+\frac{1}{\beta V_c}\partial_{u_j}\pi\right].
\]
For \(\pi_{\mathrm{eq}}(u)\propto\exp[-\beta V_c\mathcal F_N(u)]\), the bracket vanishes. Under    normalisability  and zero probability flux, this density is stationary. Omitting the state-divergence term leaves an uncancelled derivative of \(K_N\) when the dissipation depends on the field.

\subsection{{Estimation from field increments}}
\label{subsec:si_increment_estimation}

The observed increments are $\Delta z=z(t+\Delta t)-z(t)$.

For a model increment $\Delta u=u(t+\Delta t)-u(t)$, the orders in the short-time conditional moments follow from the diffusion generator. Write the coordinate equation as    \(du=f(u)dt+\sqrt{2/(\beta V_c)}B_N(u)dW_t^{(N)}\) , where \(f\) is the conditional drift and    \(B_NB_N^{\mathsf T}=K_N\) . For a smooth scalar test function \(h\colon\mathbb R^{Nq}\to\mathbb R\), its generator is
\[
\mathcal L h
=f\cdot\nabla h
+\frac{1}{\beta V_c}K_N:\nabla^2h.
\]
{Here \(\nabla\) and \(\nabla^2\) are the coordinate gradient and Hessian, and \(A:C=\sum_{i,j}A_{ij}C_{ij}\) denotes matrix contraction. For smooth coefficients, the Markov semigroup gives \(
\mathbb E[h(u(t+\Delta t))\mid u(t)=u]
=h(u)+\Delta t\,\mathcal Lh(u)+O(\Delta t^2)
\)~\cite{Pavliotis2014}. Applying this expansion to \(h_i(v)=v_i-u_i\) gives the conditional mean below. Applying it to \(h_{ij}(v)=(v_i-u_i)(v_j-u_j)\) gives the raw second moment    \(2[K_N]_{ij}\Delta t/(\beta V_c)+O(\Delta t^2)\) . Subtracting the outer product of the conditional mean changes only the \(O(\Delta t^2)\) term. Hence}
\[
\begin{aligned}
\mathbb E[\Delta u\mid u]
&=f(u)\Delta t+O(\Delta t^2),\\
\operatorname{Cov}[\Delta u\mid u]
&=\frac{2\Delta t}{\beta V_c}K_N(u)+O(\Delta t^2).
\end{aligned}
\]
{At the finite sampling interval, these relations define the short-time Gaussian quasi-likelihood used for estimation. Stage 1 estimates the conditional drift from finite differences of the observations, as specified in the system-specific implementation sections below, and gives a residual increment \(r=\Delta z-\widehat f(z)\Delta t\). Stage 2 estimates the dissipation factor from the covariance of \(r\). Let $\mathcal W$ index the magnetisation components in a group of neighbouring coarse cells, as $W$ does in Materials and Methods. Stage 2    minimises }
\[
\begin{aligned}
\mathcal L_K
&=\frac{1}{2|\mathcal W|}\Big\langle
\log\det(\Sigma_{\mathcal W}+\lambda I_{\mathcal W})
+r_{\mathcal W}^{\mathsf T}
(\Sigma_{\mathcal W}+\lambda I_{\mathcal W})^{-1}r_{\mathcal W}
\Big\rangle,\\
\Sigma_{\mathcal W}(z)
&=\frac{2\Delta t}{\beta V_c}[K_N(z)]_{\mathcal W}.
\end{aligned}
\]
Here \([K_N(z)]_{\mathcal W}\) is the principal covariance block of the complete coordinate matrix. The brackets denote an empirical average over the sampled increments and windows. The term \(\lambda I_{\mathcal W}\), with \(\lambda>0\), is a numerical covariance floor and is not part of the fluctuation--dissipation covariance. {The vector \(r_{\mathcal W}\) is the residual restricted to these components, \(I_{\mathcal W}\) is the identity on its scalar field entries, and \(|\mathcal W|\) is their number.} This covariance block includes all latent noise sources whose factor support reaches the observation window. After Stage 2, the fitted factor parameters are held fixed while Stage 3 jointly    optimises  the free energy and interaction against the observed field changes. The fitted operator $K_N(z)$ still varies with the field state $z$.

For $z^n=z(t_n)$, $\Delta z^n=z(t_{n+1})-z(t_n)$, and $\Delta t_n=t_{n+1}-t_n$, Stage 3 uses the direct drift residual
\[
\begin{aligned}
e_{F,J,i}^n
&=\left[\frac{\Delta z^n}{\Delta t_n}\right]_i
+\left[K_N(z^n)\nabla_{z}\mathcal F_N(z^n)\right]_i
-\frac{1}{\beta V_c}
\left[\operatorname{Div}_{z}K_N(z^n)\right]_i,\\
\mathcal L_{F,J}^{\mathrm{dir}}
&=\mathbb E_{\mathrm{train}}\!\left[\left|e_{F,J,i}^n\right|^2\right].
\end{aligned}
\]
Here $i$ indexes coarse-cell magnetisation components, and $\mathbb E_{\mathrm{train}}$ averages over the sampled coordinates and increments. The free-energy gradient and the state derivatives in $\operatorname{Div}_{z}K_N$ are evaluated by automatic differentiation.

%% file: SI/sections/02_ising.tex
\section{{ Ising--Kac implementation and experiments}}

\subsection{{ Microscopic dynamics and coarse graining }}

The microscopic reference is a two-dimensional periodic Ising--Kac model with heat-bath Glauber spin flips.    The microscopic index set is \(\mathcal I=(\mathbb Z/L\mathbb Z)^2\), as in the main text, and \(\mathbf r_i=i/L\in\mathbb T^2\) is the normalised microscopic  position of site \(i\). A spin \(\sigma_i\in\{-1,1\}\) at site \(i\) flips with rate
\[
c_i(\sigma)=\frac{\alpha}{2}\left[1-\sigma_i\tanh\!\left(\beta h_i^{\mathrm{loc}}(\sigma)\right)\right],
\]
where \(\alpha\) is the attempt rate and \(h_i^{\mathrm{loc}}\) is its interaction field. The local field is
\[
h_i^{\mathrm{loc}}(\sigma)
=h_{\mathrm{ext}}+\sum_{j\ne i}J(i-j)\sigma_j,
\]
where \(h_{\mathrm{ext}}\) is the external field and \(J(i-j)\) is the    normalised  lattice weight associated with the displacement \((i-j)/L\). This dynamics satisfies detailed balance with respect to the microscopic Gibbs measure. For a cell    \(\mathcal C_i=\mathcal C(\mathbf x_i)\)  containing \(V_c\) spins, the coarse    observation at $\mathbf x_i$ is \(z_i=m_i=V_c^{-1}\sum_{j\in\mathcal C(\mathbf x_i)}\sigma_j\) . The main calculations use zero external field and temperatures \(T=0.75\) and \(T=1.5\), with \(k_{\mathrm B}=1\). They use \(V_c=1024\), mapping a \(4096\times4096\) microscopic lattice to a \(128\times128\) mesoscopic field.

{The lattice spacing is \(1/4096\) in    normalised  coordinates. The Kac interaction range is \(\epsilon_{\mathrm K}=1/64\), and a block width of 32 gives \(V_c=1024\). The simulator uses adaptive \(\tau\) leaping. }

{At step \(n\), the simulator sets
\(\tau_n=\epsilon_\tau/r_{\max}(\sigma^n)\), where
\(\epsilon_\tau=0.005\) and \(r_{\max}(\sigma^n)\) is the largest spin-flip rate in the current microscopic configuration. One microscopic configuration is recorded after each \(\tau\)-leap step and is subsequently block averaged. The stored times therefore define the nonuniform intervals \(\Delta t_n\) used to form the forward increments.}

\paragraph*{Gaussian random field    initialisation .}
We generate two independent mean-zero Gaussian random fields \(\varphi_{\mathrm L},\varphi_{\mathrm S}\colon\mathbb T^2\to\mathbb R\) with Gaussian power spectra. Their correlation lengths are \(1/16\) and \(3/512\) in the unit-torus coordinates, and they are rescaled to standard deviations 1.5 and 2.0. The initial    magnetisation-bias  field is
\[
m_0(\mathbf x)
=\tanh\!\left(
\frac{\varphi_{\mathrm L}(\mathbf x)+0.4\varphi_{\mathrm S}(\mathbf x)}{0.2}
\right).
\]
At microscopic position    \(\mathbf r_\ell\) , the initial spin is sampled independently with    \(\Pr[\sigma_\ell(0)=+1]=(1+m_0(\mathbf r_\ell))/2\) . The initial coarse field is then obtained by the prescribed block average.
\subsection{{Independent theoretical references}}

The hydrodynamic reference is written on the unit torus. Let \(J\colon\mathbb T^2\to\mathbb R_+\) be the periodic Gaussian Kac kernel with range \(\epsilon_{\mathrm K}>0\),    normalised  by \(\int_{\mathbb T^2}J(\mathbf x)\,d\mathbf x=1\). This continuous kernel is defined by the Fourier coefficients
\[
J^{\mathsf F}(\mathbf n)
=\exp\!\left[-\frac{\epsilon_{\mathrm K}^2}{2}
|2\pi\mathbf n|^2\right],
\qquad \mathbf n\in\mathbb Z^2.
\]
The simulator evaluates this kernel at the microscopic lattice displacements by Fourier synthesis. The resulting lattice weights have unit sum before the self-interaction coefficient is omitted from the local field. For the heat-bath rate above, the conditional spin drift is
\[
\mathbb E[d\sigma_i\mid\sigma]/dt
=-\alpha\sigma_i
+\alpha\tanh\!\left(\beta h_i^{\mathrm{loc}}(\sigma)\right).
\]
With the time unit chosen so that \(\alpha=1\), this relation gives the reference mean evolution after coarse graining. Under Kac scaling, the    interaction range in lattice spacings diverges and each averaging cell contains increasingly many spins ($V_c\to\infty$), while its diameter is small relative to that range. The lattice  interaction converges to   convolution with    $J$, and fluctuations of the block averages vanish . The block    magnetisation  then converges to
\[
{\partial_t m(\mathbf x,t)
=-m(\mathbf x,t)
+\tanh\!\left(\beta[(J*m)(\mathbf x,t)+h_{\mathrm{ext}}]\right).}
\]
{The right-hand side is denoted by \(f_{\mathrm{jump}}[m]\). The corresponding homogeneous free-energy density is}
\[
g_{\mathrm{ref}}(m)
=\frac{1}{\beta}\left[
\frac{1+m}{2}\log\frac{1+m}{2}
+\frac{1-m}{2}\log\frac{1-m}{2}
\right]-\frac{1}{2}m^2+C.
\]
The entropy terms are continuously extended to \(m=\pm1\) using \(0\log0=0\). The reference functional is
\[
\mathcal F_{\mathrm{ref}}[m]
=\int_{\mathbb T^2}
\left[g_{\mathrm{ref}}(m)-h_{\mathrm{ext}}m
+\tfrac12m\,\mathcal L_Jm\right]\,d\mathbf x.
\]
For fields uniformly bounded away from \(m=\pm1\), its functional derivative is
\[
\frac{\delta\mathcal F_{\mathrm{ref}}[m]}{\delta m}
=\beta^{-1}\operatorname{arctanh}(m)-J*m-h_{\mathrm{ext}}.
\]
The monotonicity of $\tanh$ makes the hydrodynamic drift oppose this free-energy derivative, so it can be expressed as $\partial_t m=-\mathcal K[m]\delta\mathcal F_{\mathrm{ref}}/\delta m$ with nonnegative reference dissipation.
Here \(C\) is an arbitrary additive constant. The drift and free-energy derivative are hydrodynamic-limit references. The covariance reference below describes fluctuations in finite averaging cells.

 At zero field, normalisation of $J$ makes a spatially uniform field satisfy $\partial_t m=-m+\tanh(\beta m)$. The linear growth rate at $m=0$ is $\beta-1$, which changes sign at $\beta_c=1$, or $T_c=1$. Equivalently, $g_{\mathrm{ref}}''(0)=\beta^{-1}-1=T-1$. This is the critical temperature of the hydrodynamic reference.

 \subsection{Finite-block covariance reference}

Let \(\Delta m_i\) be the increment of the block average in cell \(i\) over an infinitesimal interval \(dt\). A flip of spin \(\ell\in\mathcal C_i\) changes \(m_i\) by \(-2\sigma_\ell/V_c\). Conditional on the microscopic configuration \(\sigma\), the jump-process quadratic variation is therefore
\begin{equation}
{\mathbb E\!\left[(\Delta m_i)^2\mid\sigma\right]
=\frac{4}{V_c}\bar c_i\,dt+o(dt),\qquad
\bar c_i=\frac{1}{V_c}\sum_{\ell\in\mathcal C_i}c_{\ell}(\sigma).}
\label{eq:si_block_quadratic_variation}
\end{equation}
For the nonoverlapping Ising blocks used here, a single spin flip belongs to only one block. Hence \(\mathbb E[\Delta m_i\Delta m_j\mid\sigma]=o(dt)\) for \(i\ne j\). The noise term in \eqref{eq:si_coordinate_ito_equation} has coordinate covariance \(2K_N\,dt/(\beta V_c)\). Equating its diagonal entry with \eqref{eq:si_block_quadratic_variation}, then conditioning on \(m_i=m\), gives
\begin{equation}
{K_{\mathrm{cov}}^{(m)}(m)=2\beta\,\mathbb E\!\left[\bar c_i\mid m_i=m\right].}
\label{eq:si_kcov_reference}
\end{equation}
This is the finite-block instantaneous covariance reference used in the main-text Ising comparison. {It is diagonal in coarse-cell coordinates only in the \(dt\to0\) quadratic-variation limit. Covariances of sampled increments at a finite interval also include correlations induced by deterministic evolution.}

At a spatially uniform Kac-limit equilibrium value \(m^*\), the heat-bath identity \(\tanh(\beta h_i^{\mathrm{loc}})=m^*\) gives \(\mathbb E[c_i\mid m^*]=(1-m^{*2})/2\). \eqref{eq:si_kcov_reference} then reduces to \(K_{\mathrm{cov}}^{(m)}(m^*)=\beta(1-m^{*2})\).

\subsection{Ising implementation}

\paragraph*{{Stage 1 conditional drift.}}

For the Ising Stage 1 model, the drift at cell $i$ is represented by a pointwise MLP evaluated on $m_i$ and $(C m)_i$. Here $C$ is an auxiliary learned $15\times15$ unit-sum convolution used only in the drift regression. The MLP has three hidden layers of width 128 with Softplus activations. The MLP and $C$ are    optimised  jointly with a batch size of 256 for 2,000 epochs and 500 batches per epoch. Adam starts at $10^{-3}$ for both parameter groups, and both learning rates are halved every 500 epochs. Training samples are balanced over the    centre magnetisation  and patch mean using 10 bins.
The fitted drift is held fixed when forming the residual increments for Stage 2.

\paragraph*{{Stage 2 dissipation factor.}}

For \method{} in the Ising--Kac system, the factor network is evaluated on a $15\times15$ field patch and uses $\mathcal S_B=\{-1,0,1\}^2$. With periodic $\ell^\infty$ distance measured in coordinate spacings,    the state patch extends seven cells from its centre and the noise sources extend one cell, so the common bound in the main text is $R=7$ . Its three $3\times3$ convolutional layers use dilations 1, 2, and 4, so the    centre  readout has a $15\times15$ effective state receptive field. The first two layers have 32 output channels and tanh activations. The last layer returns nine direction values and one amplitude value. The direction vector is    normalised  and multiplied by a positive Softplus-transformed amplitude. The resulting factor coefficients are constrained to be even under global spin reversal. One row of $K_N$ couples a $5\times5$    neighbourhood .
The factor has one latent noise channel, $p=1$. The Stage 2 quasi-likelihood uses a $3\times3$ local observation window.

The Ising finite-range runs use the selected initial portions of three independent trajectories and a patch batch size of 256. Stage 2 starts from a fresh random factor and uses 10 equal bins of the    centre magnetisation  for state-balanced sampling. The factor is trained for 2,000 epochs and 100 batches per epoch. Adam starts at $5\times10^{-4}$, and the learning rate is halved every 200 epochs. The covariance floor $\lambda$ in the Stage 2 likelihood is $10^{-8}$.

\paragraph*{{Stage 3 free energy and interaction.}}
The fitted factor parameters are held fixed during Stage 3, while the operator remains state dependent. The free-energy density and interaction are    initialised  afresh and    optimised  jointly against the observed drift. For this reference calculation, \(\rho_\theta(m,\ell)=g_\theta(m)-h_{\mathrm{ext}}m+\tfrac12m\ell\), with \(\ell=\mathcal L_{\widehat J}m\). The homogeneous density \(g_\theta\) is even under spin reversal. The interaction \(\widehat J\) is represented by a symmetric, unit-sum \(15\times15\) periodic convolution kernel without an imposed radial form. Its coefficients are averaged between opposite offsets and projected to unit sum before use.

For the numerical    magnetisation  vector $m_N=(m_i)_{i=1}^{N}$, the free energy and its coordinate derivative are
\[
\begin{aligned}
\mathcal F_{\theta,\widehat J,N}(m_N)
&=\sum_i\left[g_\theta(m_i)-h_{\mathrm{ext}}m_i\right]
-\frac{1}{2}\sum_i m_i\left[(\widehat J_Nm_N)_i-m_i\right],\\
\left[\nabla_{m_N}\mathcal F_{\theta,\widehat J,N}(m_N)\right]_i
&=g_\theta'(m_i)+m_i-(\widehat J_Nm_N)_i-h_{\mathrm{ext}}.
\end{aligned}
\]
For a spatially uniform field, the interaction action vanishes. The plotted homogeneous density is $g_\theta(m)-h_{\mathrm{ext}}m$ up to an additive constant; each displayed curve is shifted by its own minimum.

Stage 3 uses batch size 256 for 400,000 Adam updates. The learning rate starts at $10^{-3}$ and is halved every 40,000 updates. The final update supplies the evaluated model.

The selected initial intervals of three independently generated trajectories are all used for training. Each trajectory supplies 500 recorded time steps at \(T=0.75\) and 100 at \(T=1.5\). There is no validation split or validation-based checkpoint selection.

A separately generated initial field is used for evaluation, and three microscopic    realisations  from that field provide the reference rollout. Three training seeds are evaluated at each temperature.

\subsection{{Baselines and displayed quantities}}

{The hydrodynamic baseline integrates the nonlocal limiting equation given above. The Allen--Cahn baseline replaces the Kac convolution by its isotropic second-moment expansion. Its coefficients are fixed by \(\beta\) and the Kac range. The patch CNN predicts the drift from a \(15\times15\) field patch without a free-energy or dissipation-operator    parametrisation .}

{For the    unit-normalised  Kac kernel and \(h_{\mathrm{ext}}=0\), the Allen--Cahn approximation used in the comparison is}
\[
{\partial_t m
=-m+\tanh\!\left(\beta m+\kappa\Delta m\right),
\qquad
\kappa=\frac{\beta\epsilon_{\mathrm K}^2}{2}.}
\]
{The baselines and \method{} are evaluated at the microscopic recording times. The patch CNN is trained on the same trajectories and temperature-dependent intervals as \method{}. It is trained independently for three random seeds, whose mean and population standard deviation are reported.}

{Let \(\mathbf 1\) be the coordinate vector whose entries are all one. The dissipation curve in the main figure is}
\[
{K_{\mathbf 1}^{(m)}(m)
=\mathbb E\!\left(
[K_N(u)\mathbf 1]_{i}
\,\middle|\,
u_i=m
\right).}
\]
{The learned operator and the microscopic finite-block reference are evaluated on the same nonuniform fields and conditioned on the same    magnetisation  bins. For the instantaneous spin-flip process, distinct cells have zero cross quadratic variation. For the learned finite-range operator, this response contains the on-site entry and the signed sum of all represented off-site entries in the row.}

{For the interaction plot, the learned interaction coefficients and the continuous Gaussian reference are compared on the same \(15\times15\) set of offsets. The reference coefficients are obtained by evaluating \(J\) at \(\mathbf r/N_c\), where \(N_c=128\) is the number of coarse coordinates along one side of the unit domain, and    normalising  the sampled coefficients to unit sum. The displayed curves average coefficients over offsets with the same radial distance. The    centre  coefficient is omitted without    renormalising  the remaining coefficients. This averaging is used only for    visualisation  and does not impose radial symmetry during training.}

\subsection{{Evaluation field rollouts}}

{The rollouts use the models from the main Ising figure and begin from a separately generated coarse field. The microscopic reference is the mean of three independent    realisations  from this field. The displayed \method{} evolution is deterministic. Its Wiener increment is set to zero, while the state-divergence term remains in the drift. No clipping or field postprocessing is applied.}

\begin{figure}[!htb]
\centering
\includegraphics[width=\textwidth]{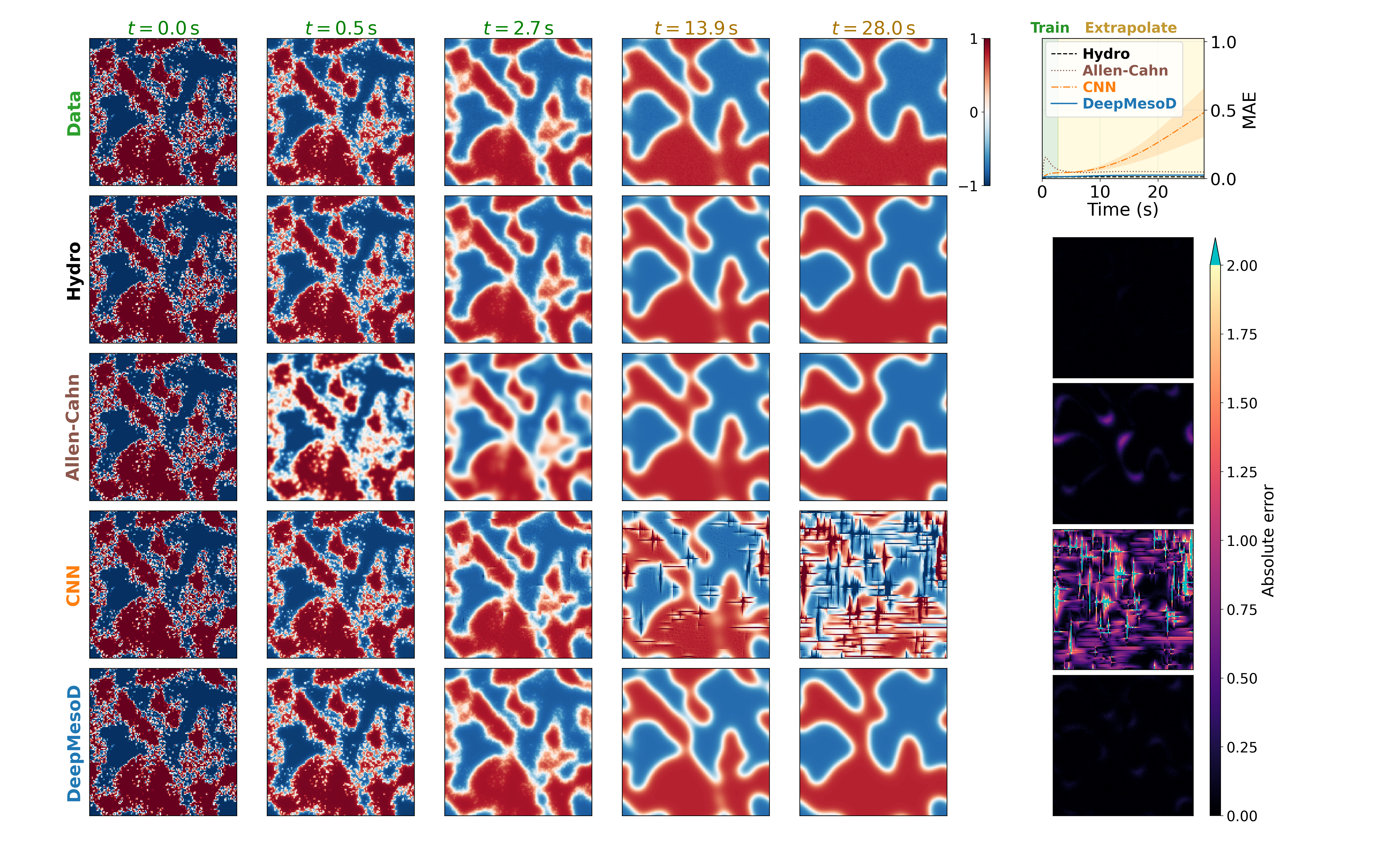}
\caption{{\textbf{Ising field evolution at \(T=0.75\).} The \method{} rollouts are noise-free with the It\^o thermal drift retained. Rows show the microscopic reference, the hydrodynamic-limit equation, the Allen--Cahn approximation, the patch CNN, and the final finite-range \method{} model. All methods start from the same evaluation field, and the microscopic reference is the average of three independent    realisations initialised  from that field. Green column labels mark saved times inside the training duration. Ochre labels mark later rollout times. Every panel uses the display range \([-1,1]\). Values outside this range in the two late CNN panels appear saturated.}}
\label{fig:si_ising_field_rollouts_t075}
\end{figure}

\begin{figure}[!htb]
\centering
\includegraphics[width=\textwidth]{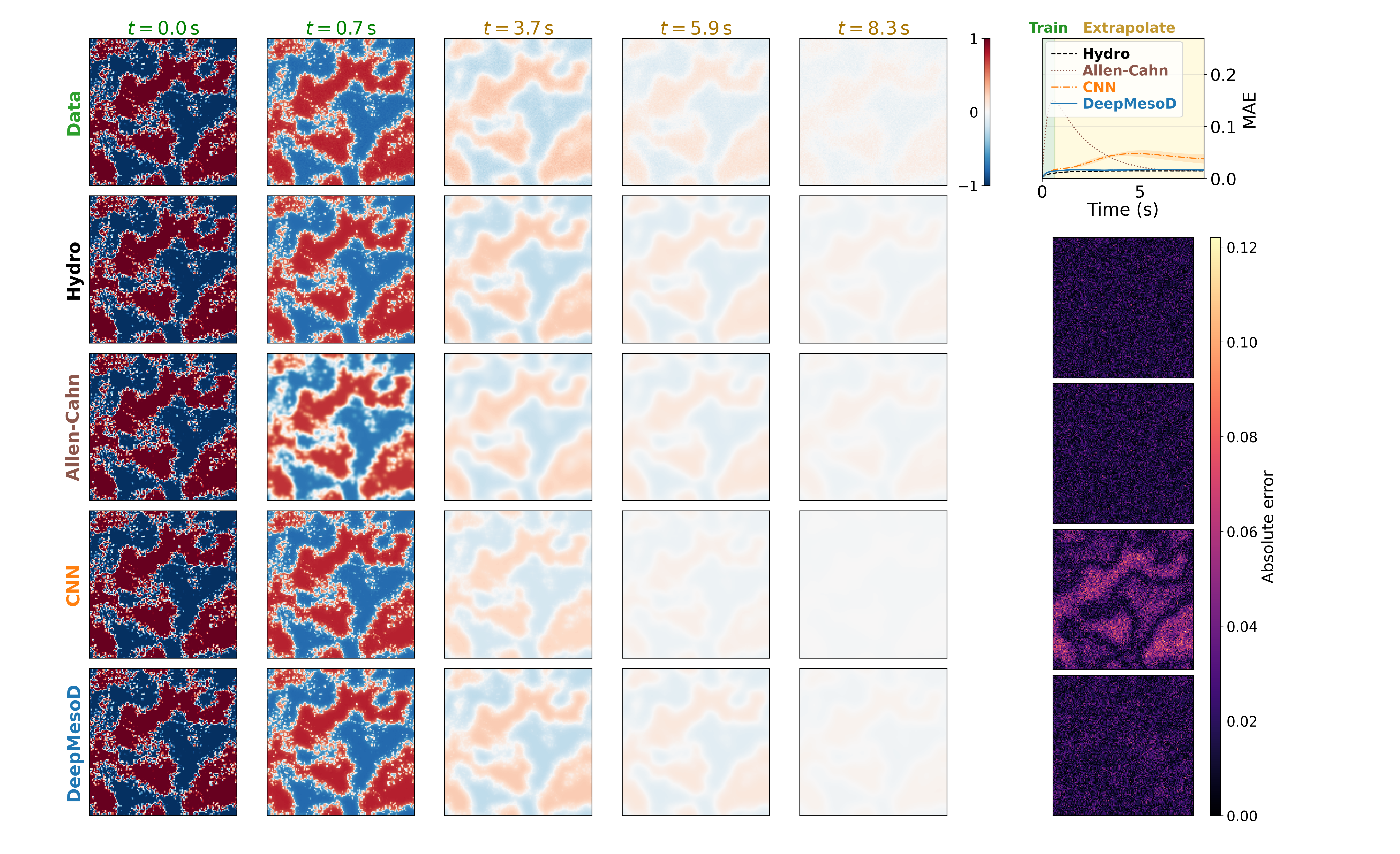}
\caption{{\textbf{Ising field evolution at \(T=1.5\).} The \method{} rollouts are noise-free with the It\^o thermal drift retained. Rows show the microscopic reference, the hydrodynamic-limit equation, the Allen--Cahn approximation, the patch CNN, and the final finite-range \method{} model. All methods start from the same evaluation field, and the microscopic reference is the average of three independent    realisations initialised  from that field. Green column labels mark saved times inside the training duration. Ochre labels mark later rollout times. Every panel uses the display range \([-1,1]\).}}
\label{fig:si_ising_field_rollouts_t15}
\end{figure}

\subsection{{Training-horizon and larger-domain evaluations}}

\begin{figure}[!htb]
\centering
\includegraphics[width=\textwidth]{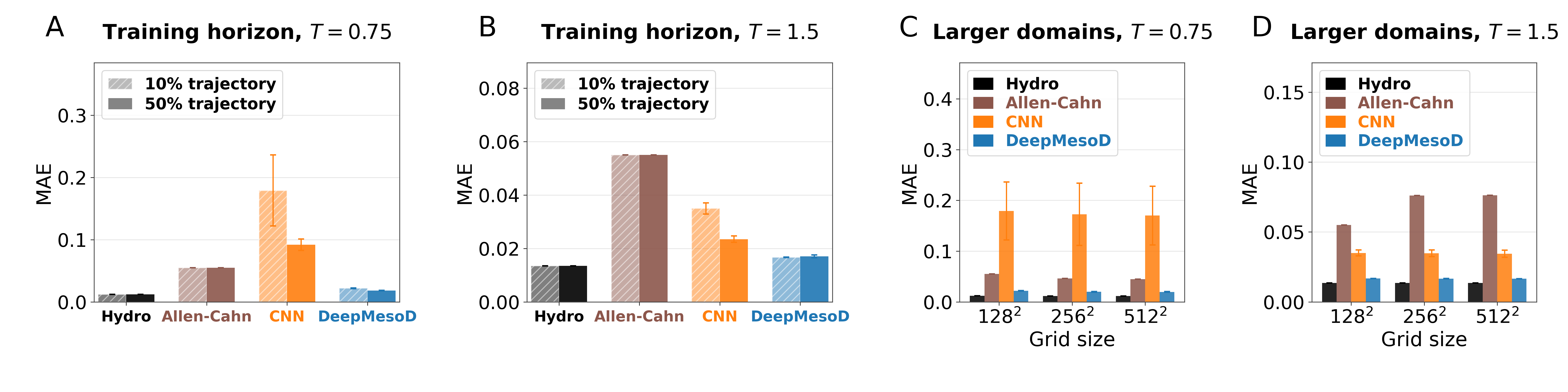}
\caption{{\textbf{Training-horizon and larger-domain diagnostics for the Stage 3 drift fit.} The \method{} rollouts are noise-free with the It\^o thermal drift retained.} {The first two panels compare the hydrodynamic-limit equation, the Allen--Cahn approximation, the patch CNN, and \method{} when the initial 10\% or 50\% of the recorded fields in each training trajectory is used.} {The last two panels compare the same methods on \(128\times128\), \(256\times256\), and \(512\times512\) periodic fields.} {The models are not retrained for the larger domains, and the mesoscopic cell scale and \(V_c\) remain fixed.} {Bars report the trajectory-averaged spatial MAE. Error bars for the learned models show one population standard deviation across three random seeds.}}
\label{fig:si_ising_larger_domains}
\end{figure}

{\Cref{fig:si_ising_larger_domains} compares rollout errors across training durations and domain sizes.} {The first two panels compare models trained with the initial 10\% and 50\% of the recorded fields in each trajectory.} {The last two panels evaluate models trained on \(128\times128\) fields without retraining on larger periodic domains.} The mesoscopic cell scale, coarse-graining volume \(V_c\), microscopic lattice spacing, and physical parameters are unchanged. For each setting, the reported quantity is the mean over the full rollout of the spatial field MAE against the corresponding microscopic reference.

At the fixed coarse-graining scale, the learned model retains comparable
rollout errors on the larger periodic domains without retraining.
Increasing the training portion from 10\% to 50\% slightly reduces the
DeepMesoD error at \(T=0.75\) and leaves it nearly unchanged at \(T=1.5\).

\subsection{{Spatial structure of the dissipation operator}}

\paragraph*{Spatial coefficient magnitudes.}

{Let \(\mathcal D=\mathcal S_B-\mathcal S_B\) be the set of displacements induced by the finite-range factor. Let \(s\in\{1,2,3\}\) index the independently trained Stage 2 factors, and write \(K_N^{(s)}\) for the coordinate dissipation matrix produced by factor \(s\). For each displacement \(\mathbf r\in\mathcal D\), we calculate the root-mean-square coordinate coefficient \(K_{\mathrm{RMS}}(\mathbf r)\)}
\[
K_{\mathrm{RMS}}(\mathbf r)
=\left(
\mathbb E_{s,u,i}\!\left[\bigl[K_N^{(s)}(u)\bigr]_{i,i+\mathbf r}^{2}\right]
\right)^{1/2}.
\]
{The expectation is the uniform empirical average over the three factor seeds, three independently generated evaluation trajectories, 24 sampled times from each trajectory, and every periodic cell index.}
At both temperatures, the RMS profile is dominated by the on-site
coefficient, while all represented off-site offsets are small on the common
   colour  scale.

\begin{figure}[!htb]
\centering
\includegraphics[width=0.72\textwidth]{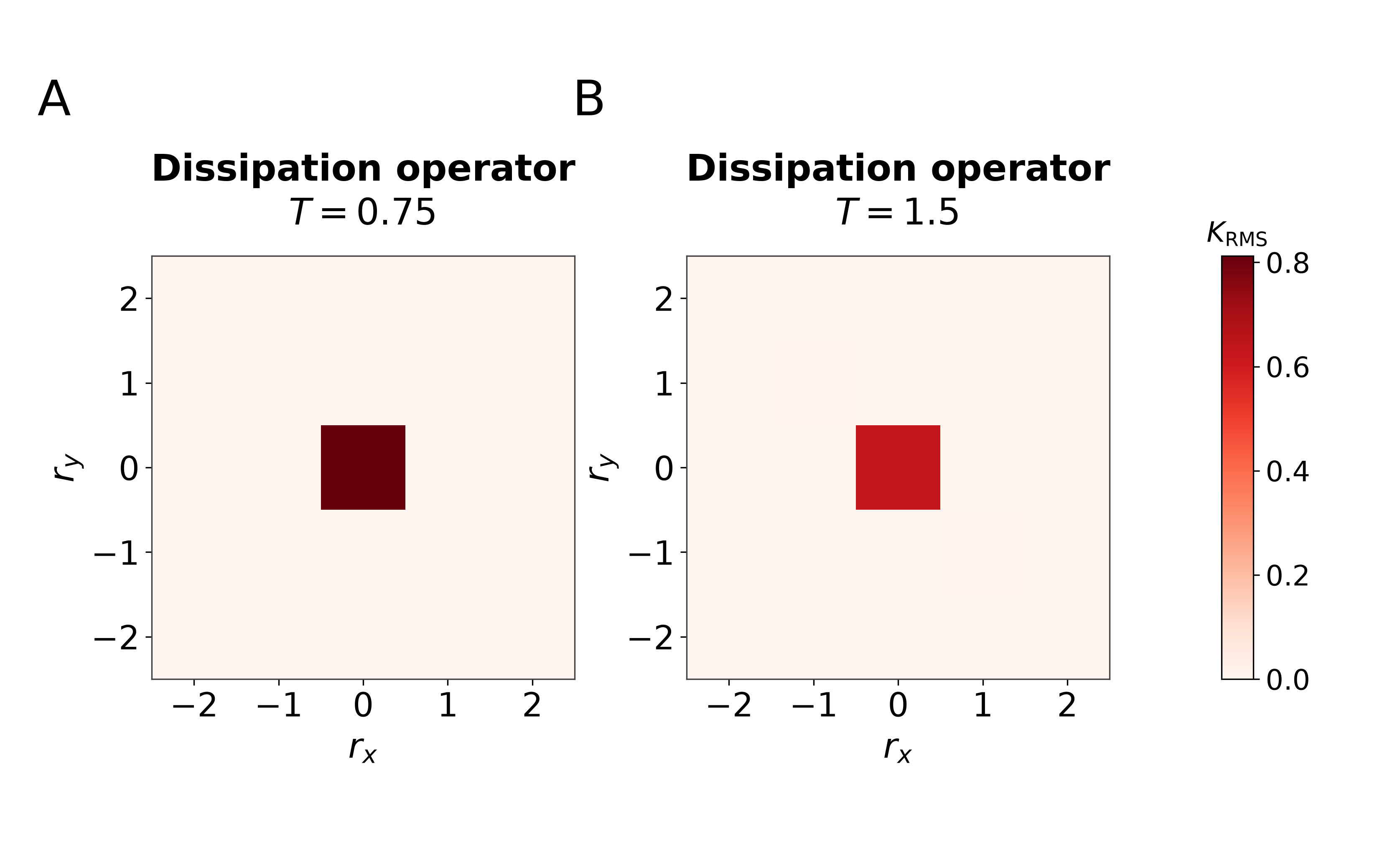}
\caption{{\textbf{Spatial structure of the learned Ising dissipation operator.}} {The plots show \(K_{\mathrm{RMS}}(\mathbf r)\), computed from the coordinate    realisation  \(K_N(u)=B_N(u)B_N(u)^{\mathsf T}\) on evaluation fields. The RMS statistic reports coefficient magnitude without sign cancellation. The displayed \(5\times5\) window is the complete coordinate support of \(K_N\). The coefficient network uses a \(15\times15\) field patch to determine these entries.}}
\label{fig:si_ising_operator_energy_diagnostics}
\end{figure}

%% file: SI/sections/03_atomistic_spin.tex
\section{Two-dimensional ferromagnet}

\subsection{Microscopic dynamics and coarse graining}
The two-dimensional ferromagnet is simulated by atomistic spin dynamics.
The same microscopic dynamical law is used for the layered
CrBr\textsubscript{3} system described in the following section. At each
magnetic site \(i\), a classical unit spin
\(\mathbf{s}_i\in\mathbb S^2\) carries magnetic moment \(\mu_s\). The general
microscopic Hamiltonian is
\[
\mathcal H(\mathbf{s})
=-\sum_{i<j}J_{ij}\,\mathbf{s}_i\mathbin{\cdot}\mathbf{s}_j
-k_u\sum_i(\mathbf{s}_i\mathbin{\cdot}\mathbf e)^2
-\mu_s\sum_i\mathbf{s}_i\mathbin{\cdot}\mathbf B
+\mathcal H_{\mathrm{dip}},
\]
where \(J_{ij}\) denotes the exchange coupling, \(k_u\) is the uniaxial
anisotropy about the easy axis \(\mathbf e\), \(\mathbf B\) is the applied
field, and \(\mathcal H_{\mathrm{dip}}\) is the dipolar interaction. The two
systems differ in their lattice geometry and material parameters. In the
CrBr\textsubscript{3} system, the stacking fault additionally changes the
exchange coupling across one interface.

Each spin evolves according to the stochastic Landau--Lifshitz--Gilbert
equation
\[
\begin{aligned}
\frac{d\mathbf{s}_i}{dt}
&=-\frac{\gamma}{1+\alpha^2}\,
\mathbf{s}_i\times
\left[\mathbf{H}_i
+\alpha\,\mathbf{s}_i\times\mathbf{H}_i\right],\\
\mathbf{H}_i
&=-\frac{1}{\mu_s}\frac{\partial\mathcal H}{\partial\mathbf{s}_i}
+\boldsymbol{\zeta}_i.
\end{aligned}
\]
Here \(\gamma\) is the gyromagnetic ratio, \(\alpha\) is the Gilbert damping,
and \(\boldsymbol{\zeta}_i\) is a zero-mean thermal field whose covariance is
fixed by the microscopic fluctuation--dissipation relation at temperature
\(T\). The precessional and damping terms preserve
\(\lvert\mathbf{s}_i\rvert=1\). We integrate this dynamics with
VAMPIRE v7.0.0~\cite{evans2014atomistic} using the Heun scheme. For either system,
the mesoscopic field in a cell \(\mathcal C_i\) containing \(V_c\)
microscopic spins is the three-component block    magnetisation
 \[
\mathbf{m}_i(t)=\frac{1}{V_c}
\sum_{j\in\mathcal C_i}\mathbf{s}_j(t)\in\mathbb R^3,
\qquad \lvert\mathbf{m}_i(t)\rvert\leq1.
\]

\paragraph*{System and coarse-graining parameters.}
The two-dimensional reference system    specialises  the Hamiltonian above to
   nearest-neighbour  ferromagnetic exchange and easy-axis anisotropy,
\[
\mathcal H_{\mathrm{2D}}(\mathbf{s})
=-\frac{1}{2}\sum_{\langle i,j\rangle}
J\,\mathbf{s}_i\mathbin{\cdot}\mathbf{s}_j
-K_u\sum_i(s_i^z)^2.
\]
The parameters are \(J=1.0\times10^{-21}\,\mathrm{J}\),
\(K_u=5.0\times10^{-23}\,\mathrm{J/atom}\), and
\(\mu_s=3.0\,\mu_B\), following Ref.~\cite{jenkins2022breaking}. The Gilbert
damping is \(\alpha=1\), and the applied field is zero. The system is a
\(1024\times1024\) monolayer at temperature
\(T=10\,\mathrm{K}\). The microscopic integration step is
\(0.1\,\mathrm{fs}\), and the trajectory is sampled at \(1\,\mathrm{ps}\)
intervals. Ten independent trajectories are    initialised  from random spin
configurations and evolved for \(2.5\times10^6\) integration steps
(\(250\,\mathrm{ps}\)). Frames from \(75\) to
\(150\,\mathrm{ps}\) are used to estimate the drift and free energy, the
first \(150\,\mathrm{ps}\) are used to estimate the dissipation, and the
\(150\)--\(250\,\mathrm{ps}\) segments of these same trajectories are reserved for later-time evaluation.
Nonoverlapping \(16\times16\) blocks give \(V_c=256\) and map the microscopic
lattice to a \(64\times64\) mesoscopic vector field. Consecutive saved fields
provide the \(1\,\mathrm{ps}\) forward increments used in estimation.

\subsection{Assessment of reversible dynamics under coarse graining}
\label{subsec:reversible_dynamics_coarse_graining}
In the present work, we restrict attention to the easy-axis    magnetisation
 \(m_z\) and represent its coarse-grained evolution by Model-A dynamics. The
microscopic stochastic LLG equation also contains reversible precession, so
this representation requires that reversible motion make no appreciable
contribution to the coarse-grained \(m_z\) channel. Our claim is that block
averaging suppresses spatially incoherent transverse precession, while any
coherent precession that remains is a rotation about the easy axis and hence
changes \((m_x,m_y)\) without changing \(m_z\).

To test this claim, we measure four complementary quantities: the transverse
angular velocity, antisymmetric cross-correlations between    magnetisation
 components, the dependence of high-frequency transverse spin-wave power on
the coarse-graining volume \(V_c\), and the circular    polarisation  and
frequency of the surviving transverse motion. Together, these quantities
test both parts of the claim: whether incoherent precession is suppressed by
coarse graining and whether the remaining reversible motion couples to
\(m_z\).

Reversible rotation in the transverse \((m_x,m_y)\) plane is measured by the
domain-sign-weighted angular velocity
\[
\omega_{\mathrm{prec}}
=
\frac{
\left\langle
\operatorname{sign}\!\left(\langle m_{z,i}\rangle_t\right)
\left(m_{x,i}\,\partial_t m_{y,i}
-m_{y,i}\,\partial_t m_{x,i}\right)
\right\rangle_{i,t}
}{
\left\langle m_{x,i}^2+m_{y,i}^2\right\rangle_{i,t}
}.
\]
Here \(i\) indexes coarse cells and \(\langle\cdot\rangle_t\) denotes a
temporal average. The domain-sign factor aligns the opposite precessional
handedness of up- and    down-magnetised  domains before averaging.

More generally, we compute the antisymmetric cross-correlations
\[
J_{ab}
=
\frac{
\left\langle
\operatorname{sign}\!\left(\langle m_{z,i}\rangle_t\right)
\left(
\delta m_{a,i}\,\partial_t m_{b,i}
-\delta m_{b,i}\,\partial_t m_{a,i}
\right)
\right\rangle_{i,t}
}{\sigma_a\sigma_b},
\qquad
\delta m_{a,i}=m_{a,i}-\langle m_{a,i}\rangle_t,
\]
where \(\sigma_a\) is the standard deviation of component \(a\). The
in-plane quantity \(J_{xy}\) probes transverse rotation, while \(J_{zx}\)
and \(J_{zy}\) test whether reversible motion couples directly to the
coarse-grained \(m_z\) channel.

To determine how spatial averaging affects incoherent precession, we measure
the transverse spin-wave power over a high-frequency band,
\[
P_{\mathrm{SW}}(V_c)
=\int_{100\,\mathrm{GHz}}^{500\,\mathrm{GHz}}
\left[
\left|\widetilde m_x(f;V_c)\right|^2
+\left|\widetilde m_y(f;V_c)\right|^2
\right]df,
\]
and compare \(P_{\mathrm{SW}}(V_c)/P_{\mathrm{SW}}(1)\) across block sizes.
This isolates the suppression of spatially incoherent transverse modes by
coarse graining.

The handedness and frequency of the transverse motion that remains after
coarse graining are    characterised  by the circularly    polarised  spectrum
\[
P_{\mathrm{circ}}(f)
=\left|\widetilde m_x(f)+\mathrm i\widetilde m_y(f)\right|^2
-\left|\widetilde m_x(f)-\mathrm i\widetilde m_y(f)\right|^2
\]
and its power-weighted mean frequency
\[
\overline f_{\mathrm{prec}}
=\frac{
\int f\,\left|P_{\mathrm{circ}}(f)\right|df
}{
\int \left|P_{\mathrm{circ}}(f)\right|df
}.
\]
Together, these quantities distinguish incoherent spin-wave dephasing from
coherent rotation and determine whether either contribution produces a
reversible component in the coarse-grained \(m_z\) dynamics.

Upon spatial coarse-graining, both the high-frequency and relaxational spectral powers decrease approximately inversely with the block volume, indicating that incoherent fluctuations are progressively averaged out. A coherent, single-handed precession about the \(z\) axis nevertheless remains identifiable, with a mean frequency of \(5.42\,\mathrm{GHz}\), corresponding to an effective field of \(193.3\pm1.5\,\mathrm{mT}\). Importantly, the associated reversible coupling is confined to the transverse \(m_x\)–\(m_y\) plane: \(J_{xy}\) is finite, whereas \(J_{zx}\) and \(J_{zy}\) are statistically consistent with zero. Thus, although transverse spins precess coherently, this motion does not generate a reversible contribution to the longitudinal \(m_z\) dynamics, supporting an effective Model-A description of the \(m_z\) channel.

\begin{figure}[!htb]
\centering
\includegraphics[width=\textwidth]{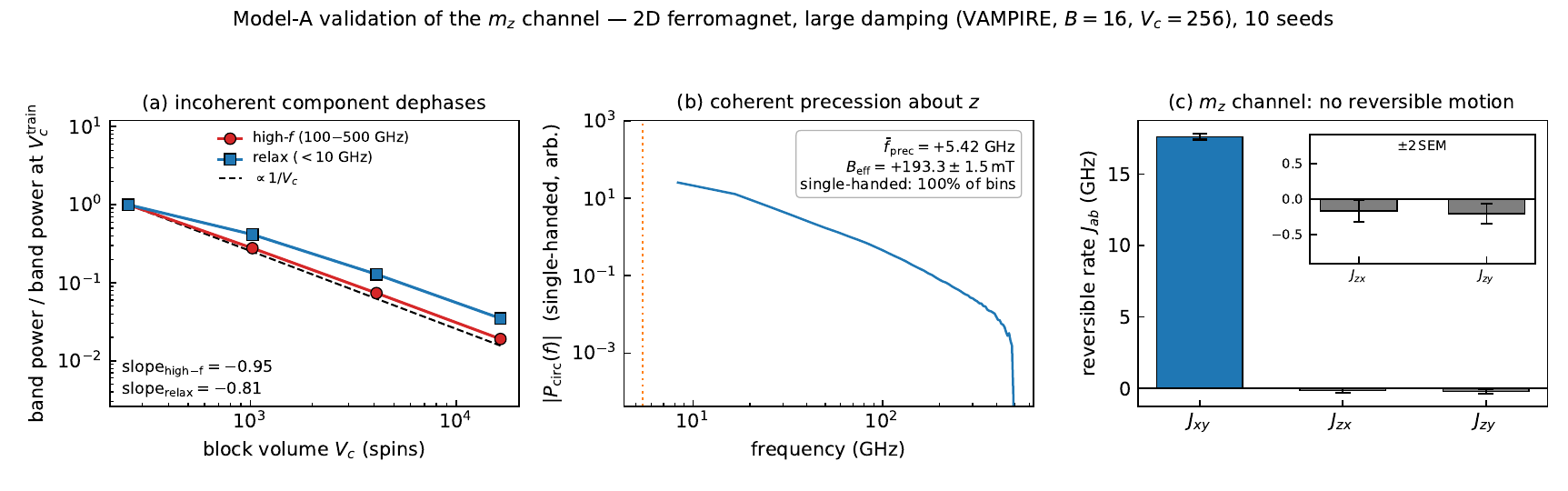}
\caption{
Validation of the effective Model-A description for the longitudinal
   magnetisation  \(m_z\) in a two-dimensional ferromagnet with large damping.
\textbf{(a)} The spectral power in both the high-frequency
(\(100\text{--}500\,\mathrm{GHz}\)) and relaxational
(\(<10\,\mathrm{GHz}\)) bands decreases with the coarse-graining volume
\(V_c\), with fitted exponents \(-0.95\) and \(-0.81\), respectively,
close to the \(V_c^{-1}\) scaling expected for incoherent fluctuations.
\textbf{(b)} The circularly    polarised  spectrum reveals a coherent, single-handed
precession about the \(z\) axis, with mean frequency
\(\bar f_{\mathrm{prec}}=+5.42\,\mathrm{GHz}\), corresponding to
\(B_{\mathrm{eff}}=193.3\pm1.5\,\mathrm{mT}\).
\textbf{(c)} The inferred reversible generator is dominated by \(J_{xy}\), which
describes rotation within the transverse \(m_x\)--\(m_y\) plane, while
\(J_{zx}\) and \(J_{zy}\) remain consistent with zero (inset, with error bars
denoting \(\pm2\) SEM). The coherent transverse precession therefore produces
no reversible motion in the \(m_z\) channel. Results are obtained from
VAMPIRE simulations with \(16\times16\) coarse-graining blocks, \(V_c=256\), and 10 independent seeds.
}
\label{fig:modelA_2D}
\end{figure}

\subsection{Implementation}

\subsubsection{Stage 1: conditional drift}
Stage 1 estimates the conditional mean of the coarse-field increments. At
cell \(i\), the drift model is
\[
\widehat{\mathbf a}_{\eta}(\mathbf m)_i
=\operatorname{MLP}_{\eta}\!\left(
\mathbf m_i,\,
(\widehat J^{(1)}\mathbf m)_i,\,
\beta,\,
h
\right),
\]
where \(\widehat J^{(1)}\) is a learned \(13\times13\) circular-convolution
kernel. The same spatial kernel is applied componentwise to \(m_x\), \(m_y\),
and \(m_z\). Thus, the pointwise input to the MLP is
\[
\left[
m_x,m_y,m_z,\,
(\widehat J^{(1)}\mathbf m)_x,
(\widehat J^{(1)}\mathbf m)_y,
(\widehat J^{(1)}\mathbf m)_z,\,
\beta,h
\right]\in\mathbb R^8.
\]
The MLP has three hidden layers of width \(64\), with \(\tanh\)
activations, and a three-dimensional output.

For two consecutive saved fields \(\mathbf m^n=\mathbf m(t_n)\) and
\(\mathbf m^{n+1}=\mathbf m(t_{n+1})\), let
\[
\mathbf d^n
=\frac{\mathbf m^{n+1}-\mathbf m^n}{\Delta t_n},
\qquad
\Delta t_n=t_{n+1}-t_n.
\]
The convolution and MLP are trained jointly by the mean-squared error between
\(\mathbf d^n\) and \(\widehat{\mathbf a}_{\eta}(\mathbf m^n)\). No
additional    regularisation  term is used. Stage 1 is used only to estimate the
conditional mean that is subtracted from the increments in Stage 2. Neither
its MLP nor its auxiliary kernel \(\widehat J^{(1)}\) enters the final
mesoscopic equation.

\subsubsection{Stage 2: dissipation factor}
Stage 2 freezes the Stage 1 model and learns a finite-range factor
\(B_h[\mathbf m]\). The dissipation operator is defined by
\[
K_h[\mathbf m]
=B_h[\mathbf m]B_h[\mathbf m]^{\mathsf T},
\]
which makes it symmetric and positive semidefinite by construction. The
factor acts on a three-component latent field \(\boldsymbol\xi\) over the
radius-one offset set
\[
\mathcal S_B=\{-1,0,1\}^2.
\]
Its coordinate action is
\[
\bigl[B_h(\mathbf m)\boldsymbol\xi\bigr]^a_i
=\sum_{s\in\mathcal S_B}\sum_{c=1}^{3}
b_i(s;\mathbf m)^{ac}\,\xi^c_{i+s},
\qquad
b_i(s;\mathbf m)\in\mathbb R^{3\times3}.
\]
A three-layer circular CNN reads the local    magnetisation  environment and
returns one \(3\times3\) matrix for each of the nine offsets, or
\(9\times3\times3=81\) coefficients at every cell.

For a displacement \(d\), the corresponding \(3\times3\) block of the
dissipation operator is
\[
\bigl[K_h(\mathbf m)\bigr]_{i,i+d}^{ae}
=
\sum_{s\in\mathcal S_B}\sum_{c=1}^{3}
b_i(s;\mathbf m)^{ac}
b_{i+d}(s-d;\mathbf m)^{ec},
\]
where a factor coefficient is zero when its offset lies outside
\(\mathcal S_B\). Because \(B_h\) has radius one, \(K_h\) has maximum radius
two.

The covariance observation window has radius one,
\[
\mathcal W_i=i+\{-1,0,1\}^2.
\]
It therefore contains nine cells and
\(d_{\mathcal W}=3|\mathcal W|=27\) scalar field entries. Let
\(K_{h,\mathcal W}\) be the corresponding principal covariance block of the
complete periodic operator. Conditional on \(\mathbf m^n\), Stage 2 models
the short-time increment as
\[
\Delta\mathbf m_{\mathcal W}^n
\sim
\mathcal N\!\left(
\Delta t_n\widehat{\mathbf a}_{\eta,\mathcal W}(\mathbf m^n),\,
\Sigma_{\mathcal W}^n
\right),
\qquad
\Sigma_{\mathcal W}^n
=\frac{2\Delta t_n}{\beta V_c}
K_{h,\mathcal W}[\mathbf m^n].
\]
With
\[
\mathbf r_{\mathcal W}^n
=\Delta\mathbf m_{\mathcal W}^n
-\Delta t_n
\widehat{\mathbf a}_{\eta,\mathcal W}(\mathbf m^n),
\]
the factor is fitted by the full-covariance Gaussian negative
log-likelihood
\[
\mathcal L_{K}
=\frac{1}{2d_{\mathcal W}}
\mathbb E\!\left[
\log\det\Sigma_{\mathcal W}^n
+(\mathbf r_{\mathcal W}^n)^{\mathsf T}
(\Sigma_{\mathcal W}^n)^{-1}
\mathbf r_{\mathcal W}^n
\right].
\]
The    normalised  quadratic residual of the fitted model is
\[
\frac{1}{d_{\mathcal W}}
\mathbb E\!\left[
(\mathbf r_{\mathcal W}^n)^{\mathsf T}
(\Sigma_{\mathcal W}^n)^{-1}
\mathbf r_{\mathcal W}^n
\right]
\simeq1.019,
\]
close to the unit value expected when the learned covariance has the correct
noise scale.

The learned operator is strongly    localised . To quantify its spatial
structure, define the distance-binned block magnitude relative to the
on-site block by
\[
R_K(d)
=
\frac{
\mathbb E_{i,\mathbf m,\,|r|=d}
\left\|[K_h(\mathbf m)]_{i,i+r}\right\|_{\mathrm F}
}{
\mathbb E_{i,\mathbf m}
\left\|[K_h(\mathbf m)]_{i,i}\right\|_{\mathrm F}
}.
\]
The measured value is approximately \(R_K(1)=4.2\%\), while the represented
couplings at larger distances decrease rapidly to below \(0.5\%\).

\subsubsection{Stage 3: free energy and interaction}
Stage 3 freezes the fitted factor \(B_h\), and hence \(K_h\), while jointly
learning the free-energy density and the final interaction kernel. The free
energy is
\[
\mathcal F_{\theta,J}[\mathbf m]
=
\sum_y
\Psi_{\theta}\!\left(
\mathbf m_y,\,
(\widehat J^{(3)}\mathbf m)_y,\,
\beta,\,
h
\right).
\]
Here \(\widehat J^{(3)}\) is a learned \(13\times13\) circular convolution
applied componentwise to the three    magnetisation  components. It is    initialised
 and trained independently of the auxiliary Stage 1 kernel
\(\widehat J^{(1)}\). The local density \(\Psi_\theta\), the system-specific energy density corresponding to $\rho_\theta$ in the general formulation, is an MLP with three
hidden layers of width \(64\), \(\tanh\) activations, and one scalar output.

Let
\(\boldsymbol\ell=\widehat J^{(3)}\mathbf m\), and let
\(\partial_1\Psi_\theta\) and \(\partial_2\Psi_\theta\) denote derivatives
with respect to the local    magnetisation  and interaction-field inputs,
respectively. The thermodynamic force is obtained by differentiating the
complete free energy:
\[
\mathbf g_{\theta,J}(x)
=
\frac{\delta\mathcal F_{\theta,J}}{\delta\mathbf m(x)}
=
\partial_1\Psi_\theta(\mathbf m_x,\boldsymbol\ell_x,\beta,h)
+
\left[
(\widehat J^{(3)})^{\mathsf T}
\partial_2\Psi_\theta(\mathbf m,\boldsymbol\ell,\beta,h)
\right]_x.
\]
Thus, differentiation includes both the direct dependence on
\(\mathbf m_x\) and the dependence of    neighbouring  energy densities on
\(\mathbf m_x\) through the convolution.

The conditional drift of the final finite-\(V_c\) equation is
\[
\mathbf a_{\theta}(\mathbf m)
=
-K_h[\mathbf m]\mathbf g_{\theta,J}(\mathbf m)
+
\frac{1}{\beta V_c}
\operatorname{Div}_{\mathbf m}K_h[\mathbf m].
\]
Noise-free prediction integrates $\dot{\mathbf m}=\mathbf a_\theta(\mathbf m)$, retaining the thermal drift while setting the Wiener increment to zero. Writing $\varepsilon=(\beta V_c)^{-1}$, its continuous-time energy balance is
\[
\frac{d\mathcal F_{\theta,J}}{dt}
=-\mathbf g_{\theta,J}^{\mathsf T}K_h\mathbf g_{\theta,J}
+\varepsilon\mathbf g_{\theta,J}^{\mathsf T}
\operatorname{Div}_{\mathbf m}K_h.
\]
Only the first term is necessarily nonpositive. Thus, these predictions are distinct from pure gradient flow, which omits both thermal drift and noise, and from the full stochastic model, which retains both. Neither the noise-free prediction nor the full stochastic model has a general pathwise free-energy descent guarantee.

For an observed rate \(\mathbf d^n\), define
\[
\mathbf q^n
=
\frac{1}{\beta V_c}
\operatorname{Div}_{\mathbf m}K_h[\mathbf m^n]
-\mathbf d^n.
\]
The quantity
\(K_h[\mathbf m^n]\mathbf g_{\theta,J}(\mathbf m^n)-\mathbf q^n\)
is a training residual, not a samplewise physical identity: the observed
rate contains stochastic fluctuations and finite-lag error relative to the
conditional drift. Stage 3    minimises its normalised  mean-square loss
\[
\mathcal L_{\mathcal F,J}
=
\frac{
\mathbb E\!\left[
\left\|
K_h[\mathbf m^n]\mathbf g_{\theta,J}(\mathbf m^n)
-\mathbf q^n
\right\|^2
\right]
}{
\operatorname{rms}(\mathbf q)^2
}.
\]
The action of the dissipation operator is evaluated matrix-free as
\[
K_h\mathbf g_{\theta,J}
=B_h\!\left(B_h^{\mathsf T}\mathbf g_{\theta,J}\right),
\]
so the complete matrix \(K_h\) is never explicitly assembled.

\subsubsection{   Optimisation  schedule}
The three stages are trained sequentially for \(3000\), \(1000\), and
\(3000\) epochs, respectively. Every stage starts with learning rate
\(10^{-3}\). The learning rate is decayed by a factor of \(0.5\) every
\(300\) epochs in Stage 1, every \(100\) epochs in Stage 2, and every
\(300\) epochs in Stage 3. The Stage 1 checkpoint is frozen for Stage 2,
and the fitted Stage 2 factor is frozen for Stage 3. The Stage 1 auxiliary
kernel and Stage 3 interaction kernel are    initialised  and trained
independently.

\subsection{Evaluation field rollouts}

\begin{figure}[!htb]
\centering
\includegraphics[width=\textwidth]{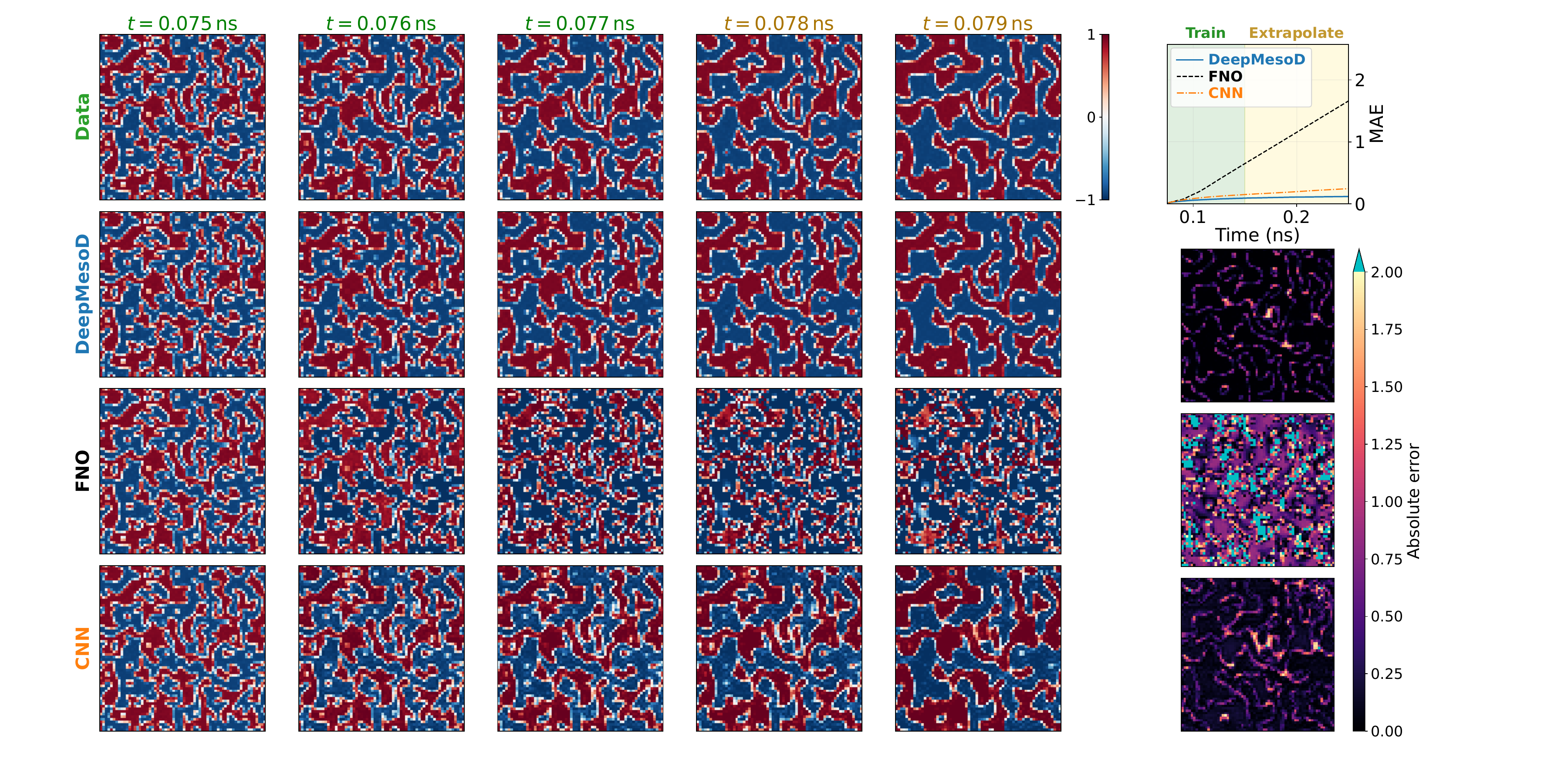}
\caption{{\textbf{Random-initial-condition rollout for the two-dimensional ferromagnet.} The DeepMesoD rollout is noise-free with the It\^o thermal drift retained. Each column shows a saved time, with green labels marking times within the training-data interval and ochre labels marking extrapolated times. Rows show the atomistic reference, the finite-range DeepMesoD model, the Fourier neural operator (FNO), and the patch CNN. The reference is the mean of independent atomistic    realisations initialised  from the same coarse field. The right-hand panels show the rollout MAE and the corresponding absolute-error fields. All    magnetisation  fields use the display range \([-1,1]\).}}
\label{fig:si_2d_random_rollout}
\end{figure}

We evaluate the learned dynamics on three qualitatively different initial
conditions. The first uses random initial spins and is therefore consistent
with the distribution represented in the training trajectories. Following a
rapid quench, the system evolves from a highly disordered state with many
small domains    towards  a state dominated by larger domains. \Cref{fig:si_2d_random_rollout}
illustrates this evolution. The drift and free-energy components are trained
using the interval from \(75\) to \(150\,\mathrm{ps}\), while DeepMesoD
remains accurate when extrapolated to \(250\,\mathrm{ps}\). In contrast, both
the FNO and the patch CNN develop substantial errors during the extrapolation
interval, with the FNO becoming unstable most rapidly.

\begin{figure}[!htb]
\centering
\includegraphics[width=\textwidth]{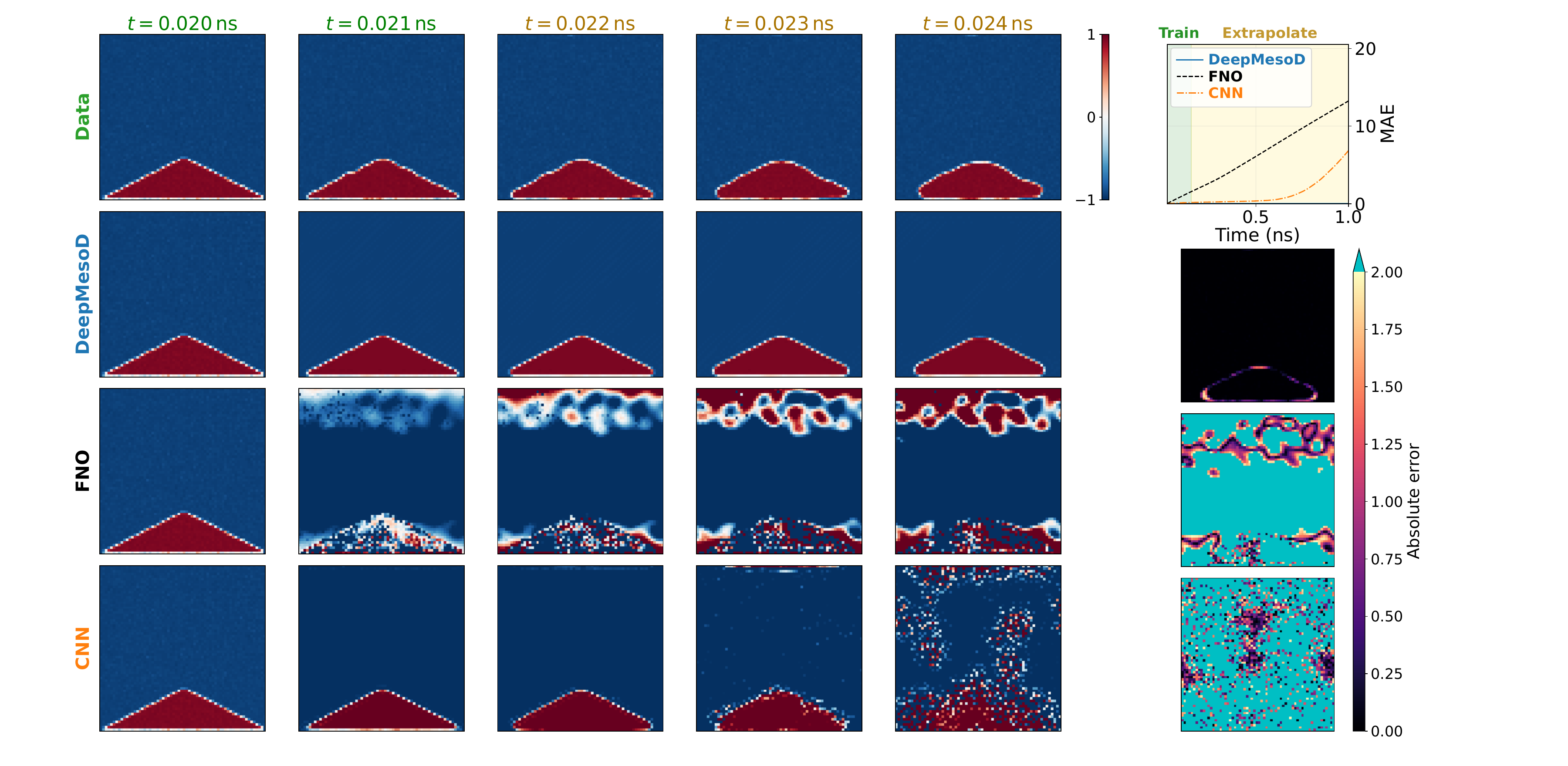}
\caption{{\textbf{Triangular-domain rollout for the two-dimensional ferromagnet.} The DeepMesoD rollout is noise-free with the It\^o thermal drift retained. The triangular minority domain is evolved from the same initial condition by the atomistic reference, DeepMesoD, the FNO, and the patch CNN. Green and ochre labels distinguish times before and after the $150\,\mathrm{ps}$ training horizon of the random-initial-state data. Neither    colour  indicates that triangular-domain data entered training. Comparisons with the atomistic reference are restricted to $0$--$1000\,\mathrm{ps}$. DeepMesoD reproduces the domain evolution over this reference interval, whereas the unconstrained baselines develop substantially larger errors. The right-hand panels show the rollout MAE and absolute-error fields only where atomistic reference data are available. All    magnetisation  fields use the display range \([-1,1]\).}}
\label{fig:si_2d_triangular_rollout}
\end{figure}

\begin{figure}[!htb]
\centering
\includegraphics[width=\textwidth]{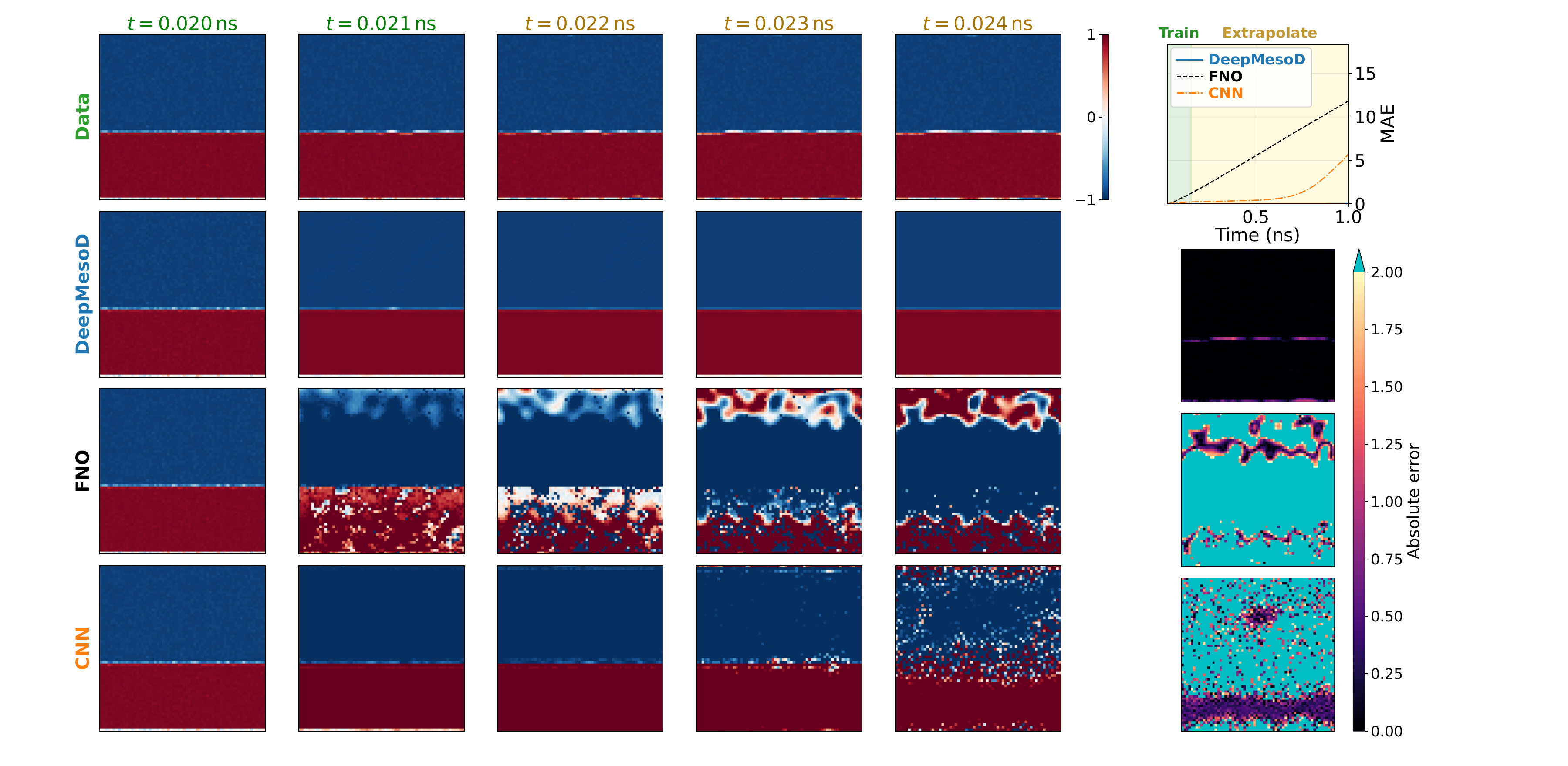}
\caption{{\textbf{Domain-wall rollout for the two-dimensional ferromagnet.} The DeepMesoD rollout is noise-free with the It\^o thermal drift retained. Time-label    colours  refer to the random-initial-state training horizon and do not indicate training on domain-wall data. The initial configuration contains a nearly planar domain wall separating oppositely    magnetised  regions. Rows compare the atomistic reference, DeepMesoD, the FNO, and the patch CNN over the observed and extrapolated time intervals. DeepMesoD maintains the position and morphology of the wall, while the baseline models accumulate larger deviations during extrapolation. The right-hand panels show the rollout MAE and absolute-error fields only where atomistic reference data are available. All    magnetisation  fields use the display range \([-1,1]\).}}
\label{fig:si_2d_domain_wall_rollout}
\end{figure}

The domain-wall initial condition tests whether the learned dynamics preserve
a coherent interface over long times. \Cref{fig:si_2d_domain_wall_rollout}
illustrates that DeepMesoD maintains the position and structure of the wall
through \(1000\,\mathrm{ps}\), whereas the two unstructured baselines
progressively diverge from the atomistic reference. The triangular-domain
initial condition provides a complementary test of relaxation    towards  the
negative, uniformly    magnetised  state. As shown in
\Cref{fig:si_2d_triangular_rollout}, DeepMesoD reproduces the expected
contraction of the minority domain and its eventual disappearance, while the
FNO and CNN again accumulate large errors during the rollout. Reference-based errors for the triangular-domain initial condition are evaluated only over $0$--$1000\,\mathrm{ps}$. The main text also shows a rollout to $25{,}000\,\mathrm{ps}$. Beyond $1000\,\mathrm{ps}$, only numerical stability, boundedness, and qualitative relaxation are assessed because no atomistic reference is available.

Together, these tests cover evolution from a disordered state, persistence of
a sharp domain wall, and curvature-driven relaxation to equilibrium. Accuracy comparisons are restricted to intervals with atomistic references. The later segments of the random-initial-state trajectories assess temporal extrapolation, while triangular-domain and domain-wall initial conditions assess geometric-initial-condition extrapolation.

%% file: SI/sections/04_crbr3.tex
\section{CrBr\textsubscript{3} film with a stacking fault}

\subsection{Microscopic dynamics and coarse graining}
The CrBr\textsubscript{3} calculations use the same classical-spin
Hamiltonian, stochastic LLG dynamics, VAMPIRE v7.0.0 Heun integrator, and
block-average definition introduced for the two-dimensional ferromagnet
above.

CrBr\textsubscript{3} is represented by its magnetic Cr\(^{3+}\) sublattice
only. Br-mediated superexchange is incorporated into effective exchange
constants. The Cr sites form a honeycomb lattice with in-plane hexagonal
lattice constant \(a=6.26\,\text{\AA}\),    nearest-neighbour  Cr--Cr distance
\[
d=\frac{a}{\sqrt{3}}=3.61\,\text{\AA},
\]
and a rectangular four-atom unit cell of dimensions
\(a\times\sqrt{3}a\). These cells are stacked with an interlayer spacing
\(c=6.12\,\text{\AA}\). Each Cr\(^{3+}\) site has \(S=3/2\), carries
\(\mu_s=3.02\,\mu_B\), and has uniaxial anisotropy
\(k_u=8.93\times10^{-24}\,\mathrm{J/atom}\) along the out-of-plane
\((001)\) easy axis. The Gilbert damping is \(\alpha=0.1\).

The exchange parameters are the intralayer    nearest-neighbour  coupling
\(J_1=2.12\times10^{-21}\,\mathrm{J}\), the ferromagnetic bulk interlayer
coupling \(J_c=3.59\times10^{-22}\,\mathrm{J/atom}\), and the
antiferromagnetic stacking-fault coupling
\(J_f=-2.77\times10^{-23}\,\mathrm{J/atom}\). The film contains \(84\)
atomic layers, corresponding to a thickness of \(51.4\,\mathrm{nm}\), with
a single antiferromagnetic stacking fault at the midplane separating two \(42\)-layer
blocks. The upper block is shifted in-plane by one    nearest-neighbour  vector,
changing the interlayer bond topology relative to the bulk stacking, and the
coupling across this interface is set to \(J_f\).
Periodic boundary conditions are imposed in \(x\) and \(y\), with lateral
sizes from \(200\) to \(800\,\mathrm{nm}\), while the two surfaces in \(z\)
are open. The \(200\times200\times51.4\,\mathrm{nm}^3\) reference film
contains approximately \(2\times10^7\) microscopic spins.

Each trajectory begins from an independently sampled random spin
configuration. At \(t=0\), the system is instantaneously quenched to
\(T=50\,\mathrm{K}\) at zero applied field and then integrated for
\(2.5\times10^5\) steps with a \(2\,\mathrm{fs}\) time step, giving a total
duration of \(500\,\mathrm{ps}\). Ten statistically independent trajectories
are generated, using a distinct random seed for both the initial spin
configuration and the thermal integrator noise in every trajectory. Full
atomic spin configurations and cell-averaged    magnetisation  fields are saved
every \(2\,\mathrm{ps}\).

For coarse graining, the Cr moments are averaged over nonoverlapping
\(5.01\times5.42\times1.84\,\mathrm{nm}^3\) cells.
Each slab spans three atomic layers, and the learning calculation uses $V_c=480$.
The in-plane cell spacings differ by \(7.6\%\), a
geometric anisotropy retained in the learned grid-index kernels. For the
\(200\times200\times51.4\,\mathrm{nm}^3\) reference film, this gives a
\(40\times37\times28\) mesoscopic vector field with \(251\) saved frames per
trajectory. Consecutive fields provide the \(2\,\mathrm{ps}\) forward
increments.

The quantities used to assess reversible motion in CrBr\textsubscript{3}
follow the definitions introduced in
\cref{subsec:reversible_dynamics_coarse_graining} for the two-dimensional
ferromagnet. The high-frequency transverse power decreases nearly inversely
with the block volume, whereas the low-frequency relaxational power is much
less sensitive to coarse graining. This contrast shows that block averaging
selectively suppresses spatially incoherent precession while retaining the
slow collective dynamics. The surviving transverse motion is single-handed
and corresponds to coherent precession about the \(z\) axis at
\(\bar f_{\mathrm{prec}}=3.53\,\mathrm{GHz}\).
The antisymmetric correlation rates \(J_{zx}\) and \(J_{zy}\) are statistically consistent with zero, while \(J_{xy}\) remains finite. These diagnostics support the dissipative description of the longitudinal    magnetisation  at the chosen coarse-graining scale.

\begin{figure}[!htb]
\centering
\includegraphics[width=\textwidth]{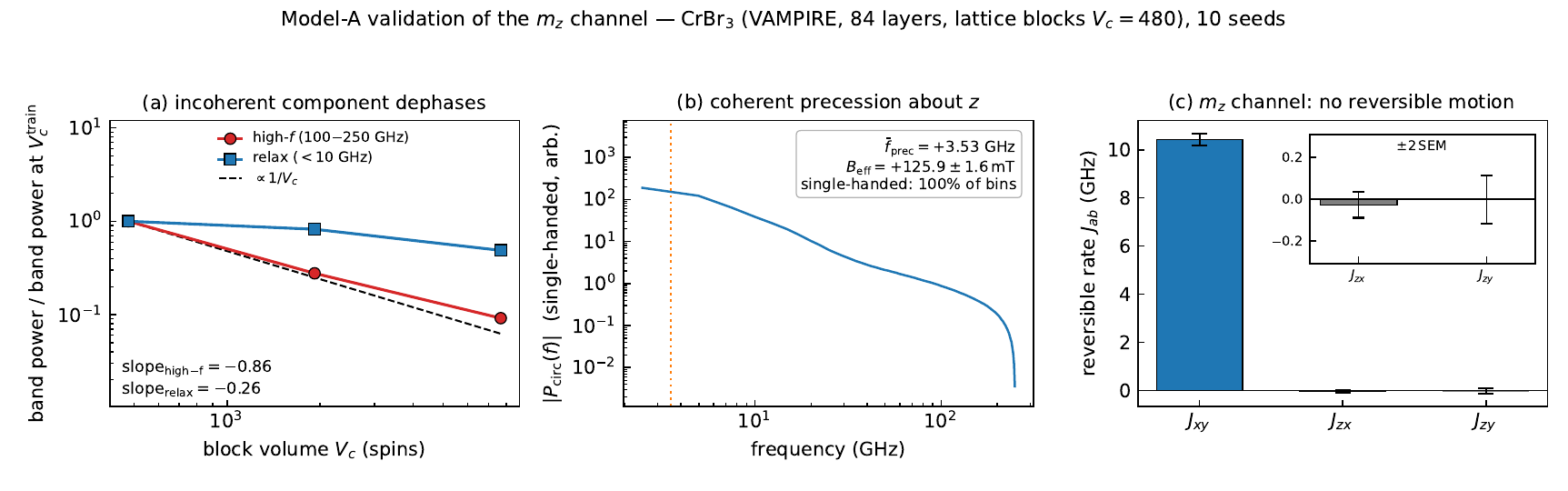}
\caption{
Validation of the effective Model-A description for the longitudinal
   magnetisation  \(m_z\) in the layered CrBr\textsubscript{3} film.
\textbf{(a)} The    normalised  high-frequency transverse spectral power
(\(100\text{--}250\,\mathrm{GHz}\)) decreases with coarse-graining volume
\(V_c\) with a fitted exponent of \(-0.86\), close to the \(V_c^{-1}\)
scaling expected for spatially incoherent fluctuations. In contrast, the
low-frequency relaxational power (\(<10\,\mathrm{GHz}\)) decreases more
slowly, with a fitted exponent of \(-0.26\).
\textbf{(b)} The circularly    polarised  spectrum identifies a coherent, single-handed
precession about the \(z\) axis, with mean frequency
\(\bar f_{\mathrm{prec}}=+3.53\,\mathrm{GHz}\), corresponding to an
effective field \(B_{\mathrm{eff}}=125.9\pm1.6\,\mathrm{mT}\).
\textbf{(c)} The antisymmetric correlation rate is dominated by \(J_{xy}\), which describes
rotation in the transverse \(m_x\)--\(m_y\) plane, whereas \(J_{zx}\) and \(J_{zy}\) are consistent with zero (inset, with error bars denoting \(\pm2\)
SEM). The coherent transverse precession therefore generates no detectable
reversible motion in the \(m_z\) channel. Results are obtained from VAMPIRE
simulations of the 84-layer film using lattice-commensurate blocks with the
training coarse-graining volume \(V_c=480\) and 10 independent seeds.
}
\label{fig:modelA_CrBr3}
\end{figure}

\subsection{Implementation}
Here $\mathbf m$ collects the coarse-cell magnetisations, and $B_N(\mathbf m)$ and $K_N(\mathbf m)$ denote the corresponding numerical factor and dissipation matrices.
The CrBr\textsubscript{3} model follows the same three-stage construction as
the two-dimensional ferromagnet: Stage 1 estimates the conditional drift,
Stage 2 learns the    factorised  dissipation operator
\(K_N=B_NB_N^{\mathsf T}\), and Stage 3 learns the free energy and interaction
with the fitted factor parameters held fixed. The principal change is the treatment
of the third spatial direction. The interaction kernel couples different
slabs and can therefore represent surfaces and the stacking fault, whereas
the finite-range dissipation factor acts independently within each slab. The
working model uses \(\beta=0.02\), \(V_c=480\), and \(h=0\) during training. The numerical time unit is ns.

\subsubsection{   Factorised  three-dimensional interaction kernel}
Let \(\mathbf x=(x,y)\) index an in-plane coarse cell and let
\(z\in\{1,\ldots,L_z\}\), with \(L_z=28\), index the coarse slabs. Slab $z$ contains atomic layers $3z-2$, $3z-1$, and $3z$ in one-based numbering. The midplane fault separates atomic layers 42 and 43, hence coarse slabs 14 and 15 in this notation, or slabs 13 and 14 in the zero-based numbering used in the main text. In Stages
1 and 3, the interaction is    parametrised  as
\[
J(z_{\mathrm{out}},z_{\mathrm{in}},r)
=J_z(z_{\mathrm{out}},z_{\mathrm{in}})J_{xy}(r),
\qquad r\in\{-3,\ldots,3\}^2,
\]
and is applied componentwise to the three    magnetisation  components:
\[
(J\mathbf m)_{\mathbf x,z}
=\sum_{z'=1}^{L_z}J_z(z,z')
\sum_{r\in\{-3,\ldots,3\}^2}
J_{xy}(r)\mathbf m_{\mathbf x+r,z'}.
\]
The shared \(7\times7\) in-plane profile \(J_{xy}\) uses circular padding in
\(x\) and \(y\) and is    normalised  to unit sum. Here $J_{xy}$ is the spatial kernel of the main text, distinct from the antisymmetric correlation rates $J_{ab}$ used above. The layer matrix
\(J_z\in\mathbb R^{28\times28}\) is symmetric but is not    normalised . The
full layer matrix allows the learned interaction to vary at the two free
surfaces and across the film midplane. Neither the location nor the coupling
of the stacking fault is supplied during training. Because the coarse cells
have unequal in-plane spacings, all kernels are learned in grid-index
coordinates and the resulting geometric anisotropy is absorbed by
\(J_{xy}\).

\subsubsection{Stage 1: conditional drift}
Stage 1 uses an independently    initialised factorised  kernel \(J^{(1)}\) and
models the conditional drift as
\[
\widehat{\mathbf a}_{\eta}(\mathbf m)_{\mathbf x,z}
=\operatorname{MLP}_{\eta}\!\left(
\mathbf m_{\mathbf x,z},
(J^{(1)}\mathbf m)_{\mathbf x,z}
\right)\in\mathbb R^3.
\]
Since \(\beta\) and \(h\) are fixed in this dataset, they are not included as
network inputs. The six input coordinates are
\([\mathbf m,J^{(1)}\mathbf m]\). The same MLP is shared over all slabs. It
has three hidden layers of width \(64\), \(\tanh\) activations, and a
three-dimensional output. Thus all explicit layer dependence enters through
the learned matrix \(J_z^{(1)}\).

The regression target is the one-frame drift estimate
\[
\mathbf d^n
=\frac{\mathbf m^{n+1}-\mathbf m^n}{\Delta t},
\qquad \Delta t=0.002\,\mathrm{ns}.
\]
The convolution and MLP are trained jointly by the mean-squared error between
\(\mathbf d^n\) and \(\widehat{\mathbf a}_{\eta}(\mathbf m^n)\).
Stage 1 is trained on frames \(50\leq n<251\), corresponding to
\(100\)--\(500\,\mathrm{ps}\), for \(3000\) epochs. Adam starts at
\(10^{-3}\), and the learning rate is multiplied by \(0.8\) every \(300\)
epochs. Stage 1 is used only to define the conditional mean of the Stage 2
residual. It does not enter the final mesoscopic equation.

\subsubsection{Stage 2: dissipation factor}
Stage 2 freezes the three-dimensional Stage 1 drift and learns the same
matrix-valued finite-range factor used for the two-dimensional ferromagnet,
but applies it independently to every \(z\) slab. With
\(\mathcal S_B=\{-1,0,1\}^2\),
\[
\bigl[B_N(\mathbf m)\boldsymbol\xi\bigr]^a_{\mathbf x,z}
=\sum_{s\in\mathcal S_B}\sum_{c=1}^{3}
b_{\mathbf x,z}(s;\mathbf m)^{ac}
\xi^c_{\mathbf x+s,z}.
\]
There is therefore no dissipative coupling between distinct slabs: the
layer-to-layer dynamics is represented by the free-energy interaction
\(J_z\), not by \(B_N\). Within a slab, each of the nine offsets carries a
full \(3\times3\) matrix. A three-layer circular CNN with channels
\(3\to32\to32\to81\), \(\tanh\) activations, and an effective
\(7\times7\) state patch predicts the \(81\) coefficients at each cell. The
diagonal entries of the central block are passed through a Softplus map to
keep them positive. The complete factor is made even under global spin
reversal,
\[
b[\mathbf m]\leftarrow
\tfrac12\bigl(b[\mathbf m]+b[-\mathbf m]\bigr),
\]
but no fourfold or dihedral in-plane symmetry is imposed, because the
lattice-commensurate coarse cells are not square. As in the two-dimensional
model, \(B_N\) has radius one and \(K_N=B_NB_N^{\mathsf T}\) has maximum
in-plane radius two.

For each sampled slab, the covariance likelihood is evaluated on a
\(3\times3\) in-plane observation window \(\mathcal W\), containing
\(d_{\mathcal W}=27\) scalar    magnetisation  entries. The complete
\(5\times5\) latent halo is retained, and
\[
K_{N,\mathcal W}=L_{\mathcal W}L_{\mathcal W}^{\mathsf T}.
\]
Here $L_{\mathcal W}$ contains the rows of $B_N$ indexed by $\mathcal W$ and all latent columns contributing to those rows.

The factor is fitted with the Gaussian quasi-likelihood in \cref{subsec:si_increment_estimation}, using the frozen Stage 1 drift to form the residual increments.
Stage 2 uses all frames \(0\leq n<251\) and is trained for \(1000\) epochs
with Adam    initialised  at \(10^{-3}\). The learning rate is halved every
\(200\) epochs. Samples are balanced over \(\lvert\mathbf m\rvert\).
Ten magnitude bins supply 20 batches of 128 patches per epoch. We set the covariance floor in \cref{subsec:si_increment_estimation} to $\lambda=10^{-8}$.

\subsubsection{Stage 3: free energy and interaction}
Stage 3 holds the fitted factor parameters fixed and learns
\[
\mathcal F_{\theta,J}[\mathbf m]
=\sum_{\mathbf x,z}
\rho_{\theta}\!\left(
\mathbf m_{\mathbf x,z},
(J^{(3)}\mathbf m)_{\mathbf x,z}
\right).
\]
The    factorised  \(7\times7\)-by-\(28\times28\) interaction
\(J^{(3)}\) is    initialised  and trained independently of \(J^{(1)}\).
The local density network has six inputs \([\mathbf m,J^{(3)}\mathbf m]\),
three hidden layers of width \(64\) with \(\tanh\) activations, and one scalar
output. No explicit layer embedding is used, so interpretable layer
dependence is carried by \(J_z^{(3)}\).
Here $\mathcal F_{\theta,J}[\mathbf m]$ is the numerical free energy $\mathcal F_N(\mathbf m)$ for this layered implementation. The second input to $\rho_\theta$ is $J^{(3)}\mathbf m$, and the    unnormalised  layer matrix carries the interlayer coupling strength.

The vector $\mathbf g_{\theta,J}$ represents the free-energy derivative in the numerical coordinates,
\[
\mathbf g_{\theta,J}
=\nabla_{\mathbf m}\mathcal F_N(\mathbf m)
=\partial_1\rho_{\theta}
+(J^{(3)})^{\mathsf T}\partial_2\rho_{\theta}.
\]
The thermodynamic force is $-\mathbf g_{\theta,J}$. The derivative is evaluated by automatic differentiation on the complete field, which
retains the adjoint contribution from every energy density coupled through
\(J^{(3)}\).

For every training field, define
\[
\mathbf q^n
=\frac{1}{\beta V_c}
\operatorname{Div}_{\mathbf m}K_N[\mathbf m^n]
-\frac{\Delta\mathbf m^n}{\Delta t}.
\]
The state divergence is evaluated by local Jacobian contraction and is
precomputed because the factor parameters are held fixed. Stage 3    minimises  the direct
residual
\[
\mathcal L_{\mathcal F,J}
=\mathbb E\left\lVert
K_N[\mathbf m^n]\mathbf g_{\theta,J}(\mathbf m^n)-\mathbf q^n
\right\rVert^2
.
\]
The operator action is computed independently in each slab as
\[
K_N\mathbf g_{\theta,J}
=B_N\!\left(B_N^{\mathsf T}\mathbf g_{\theta,J}\right),
\]
without assembling the full dissipation matrix.

Stage 3 uses frames \(50\leq n<251\), with stride three for the precomputed
state-divergence targets. It is trained for \(2000\) epochs with Adam
   initialised  at \(10^{-3}\). The learning rate is halved every \(400\)
epochs.
Each epoch contains 50 updates with four complete fields per update. The selected increments start every third frame from 100 to 496 ps, giving 67 increments per trajectory, each of duration 2 ps.

\subsubsection{Zeeman calibration}
The zero-field training data do not identify the scale of the Zeeman term.
It is therefore calibrated after the three-stage fit using
\[
\rho_{\mathrm Z}=-\alpha_h h m_z,
\qquad
h=\frac{B}{\mu_0M_s},
\qquad
\mu_0M_s=0.3393\,\mathrm T.
\]
Fitting paired \(+B\) and \(-B\) increments at \(25\) and
\(50\,\mathrm{mT}\) gives
\[
\alpha_h=0.1894
\]
for the dimensionless field \(h\), or equivalently a coefficient of
\(0.5581\,\mathrm T^{-1}\) when the field is expressed directly in tesla.
The fit uses inverse-variance weighting, with statistical uncertainty $0.0089$ on $\alpha_h$.

\subsection{Prediction dynamics}

The CrBr$_3$ trajectory rollouts integrate the finite-$V_c$ conditional drift with the It\^o thermal drift retained and the Wiener increment set to zero.

The equilibrium field responses are computed from sector weights using annealed importance sampling, as described below.

\subsection{Inference with multiple stacking faults}
\label{subsec:crbr3_multifault}
The    factorised  interaction representation also permits inference for films
containing more stacking faults than the single fault present in the training
trajectories. We retain the original \(L_z=28\) coarse slabs. Let \(f_0\)
denote the fault interface in the training geometry and let
\(\mathcal F=\{f_1,\ldots,f_{N_f}\}\) be a prescribed set of interfaces in
the same 28-slab film.
We form a background matrix \(J_z^{\mathrm{bg}}\) by replacing entries near
the original fault with the median learned coupling at the same layer
separation, estimated from interior entries away from the fault. Both the
coupling range and the    fault-neighbourhood  radius are four resolved slabs.
Within this coupling range, an entry is replaced if either layer index lies
within four slabs of \(f_0\). All other entries remain unchanged. The fault
correction is \(\Delta J_z=J_z-J_z^{\mathrm{bg}}\), and the multi-fault matrix is
{
\[
J_z^{(\mathcal F)}(a,b)
=J_z^{\mathrm{bg}}(a,b)
+\sum_{f\in\mathcal F}\Delta J_z(a-f+f_0,b-f+f_0).
\]
}
Here \(\Delta J_z\) is zero outside the film's index range. Translated
corrections add where they overlap. Placing a single fault at \(f_0\)
recovers the original learned matrix.

\begin{figure}[!htb]
\centering
\includegraphics[width=\textwidth]{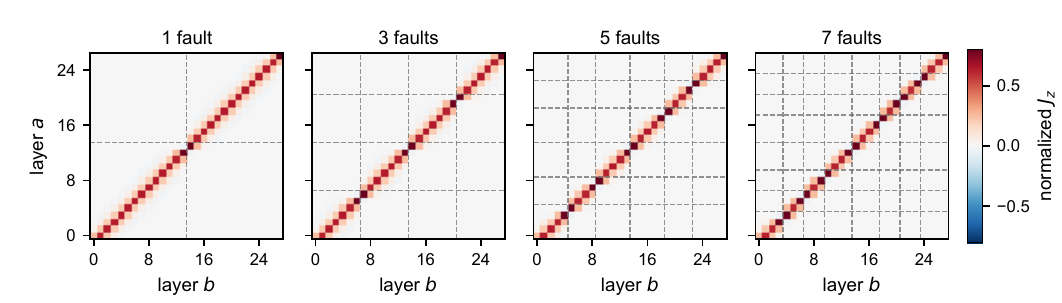}
\caption{
Construction of multi-fault interlayer kernels at fixed film thickness.
From left to right, the learned single-fault structure is used to assemble
\(28\times28\) layer matrices containing 1, 3, 5, and 7 stacking faults.
Dashed horizontal and vertical lines mark the prescribed fault interfaces.
At every added interface, the local fault correction extracted from the learned
single-fault \(J_z\) is added to the background matrix defined above. All
matrices are displayed on the same    normalised colour  scale. Only \(J_z\) is
modified. The film thickness, local free energy, in-plane interaction,
dissipation operator, and network weights remain unchanged, and no retraining
is performed.
}
\label{fig:crbr3_multifault_kernel}
\end{figure}

\Cref{fig:crbr3_multifault_kernel} shows matrices containing one,
three, five, and seven approximately evenly spaced faults. They are controlled
modifications of the geometry-dependent matrix \(J_z\). The local free-energy
density \(\rho_\theta\), in-plane kernel \(J_{xy}\), dissipation factor \(B_N\),
and all network weights remain fixed. Copying the learned fault motif to
prescribed interfaces therefore isolates how fault geometry changes the field
response within the learned model, without additional microscopic training
trajectories.

\begin{figure}[!htb]
\centering
\includegraphics[width=\textwidth]{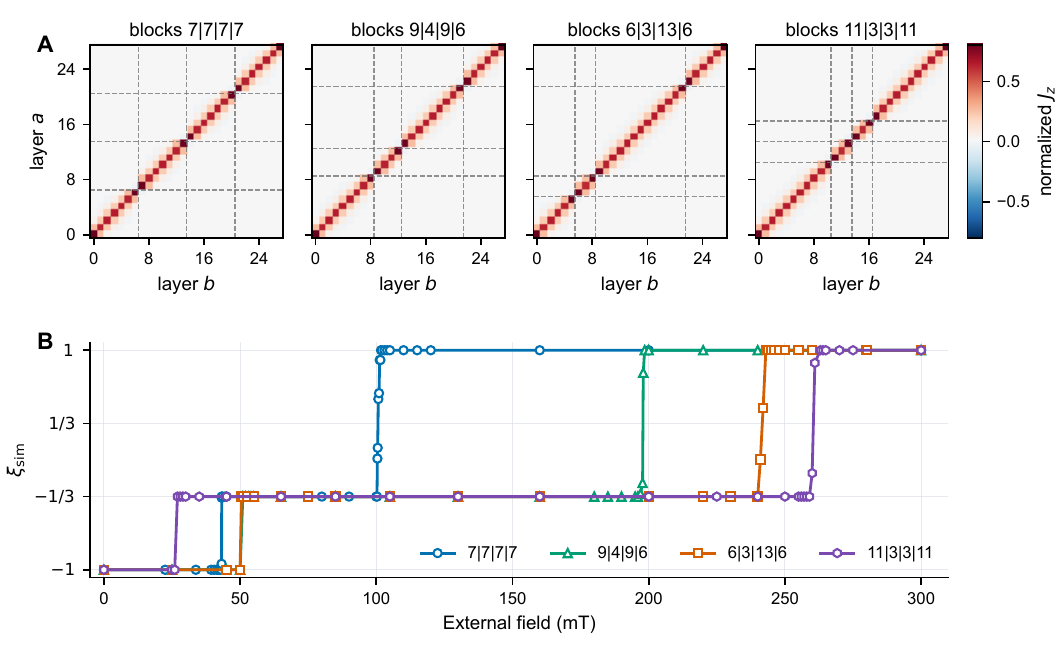}
\caption{
Effect of fault placement at fixed fault number.
\textbf{A.} Four interlayer matrices assembled from the learned fault motif,
each containing three stacking faults at different prescribed interfaces.
The labels give the corresponding block thicknesses in coarse-slab units.
\textbf{B.} Equilibrium mean fault-plane correlation
\(\xi_{\mathrm{sim}}\) as a function of external field for the same four
geometries. The field dependence is calculated from the equilibrium sector
weights described below. Points below the running maximum of
\(\xi_{\mathrm{sim}}\) as the field increases are omitted from the displayed
switching traces. The complete unfiltered values are retained in the source
data. The local free energy, in-plane interaction,
dissipation operator, field calibration, and network weights are unchanged.
}
\label{fig:crbr3_fault3_geometry_response}
\end{figure}

\subsubsection{Equilibrium sector weights from annealed importance sampling}

At a fixed dimensionless field \(h\), direct trajectories relax within a
block-alignment sector but do not visit all sectors with their equilibrium
probabilities on the simulated time scale. Each sector specifies the signs of the out-of-plane block    magnetisations . Let \(\mathcal S\) denote the set
of retained block-alignment sectors and let \(\Gamma_s\) be the configurations
in sector \(s\in\mathcal S\). For the field-dependent free energy
\(\mathcal F_h\), which abbreviates $\mathcal F_{\theta,J,h}$ of the main text for the prescribed fault geometry, the restricted partition function is
\begin{equation}
Z_s(h)
=
\int_{\Gamma_s}
\exp\!\left[-\beta V_c\mathcal F_h[\phi]\right]
\mathcal D\phi.
\label{eq:si_crbr3_restricted_partition}
\end{equation}
The integrals here are over the numerical field coordinates used in the calculations.
For a configuration \(\phi\), the fault-plane statistic is
\begin{equation}
\xi(\phi)
=
\frac{1}{N_fN_{xy}}
\sum_{f=1}^{N_f}\sum_{i=1}^{N_{xy}}
\operatorname{sign}\!\left(m_{z,i}^{f,+}\right)
\operatorname{sign}\!\left(m_{z,i}^{f,-}\right).
\label{eq:si_crbr3_fault_statistic}
\end{equation}
Here $N_f$ is the number of faults, $N_{xy}$ is the number of cells per plane, and $m_{z,i}^{f,+}$ and $m_{z,i}^{f,-}$ are the out-of-plane    magnetisations  immediately above and below fault $f$.
Its conditional equilibrium average within sector \(s\) is
\begin{equation}
\bar\xi_s(h)
=
\frac{1}{Z_s(h)}
\int_{\Gamma_s}
\xi(\phi)
\exp\!\left[-\beta V_c\mathcal F_h[\phi]\right]
\mathcal D\phi.
\label{eq:si_crbr3_sector_average}
\end{equation}
The equilibrium response reported in main-text Fig.~4E is therefore
\begin{equation}
\xi_{\mathrm{sim}}(h)
=
\sum_{s\in\mathcal S}p_s(h)\bar\xi_s(h),
\qquad
p_s(h)=\frac{Z_s(h)}{\sum_{r\in\mathcal S}Z_r(h)}.
\label{eq:si_crbr3_equilibrium_xi}
\end{equation}

Intermediate values of $\xi_{\mathrm{sim}}$ can arise from mixed alignments within a fault plane, different alignments among fault planes, or averaging over configurations.

The ratios of the restricted partition functions are evaluated by
bidirectional annealed importance sampling~\cite{neal2001annealed}. For a
bridge with    unnormalised  densities
\(\widetilde\pi_0,\ldots,\widetilde\pi_L\),    normalising  constants
\(Z_0,\ldots,Z_L\), and samples \(\phi_0,\ldots,\phi_{L-1}\) propagated by
Markov transitions that preserve the successive bridge distributions, the
forward importance weight is
\begin{equation}
w_{0\to L}
=
\prod_{\ell=1}^{L}
\frac{\widetilde\pi_\ell(\phi_{\ell-1})}{\widetilde\pi_{\ell-1}(\phi_{\ell-1})},
\qquad
\mathbb E_{0\to L}[w_{0\to L}]
=\frac{Z_L}{Z_0}.
\label{eq:si_crbr3_ais_weight}
\end{equation}
The reverse schedule estimates the reciprocal ratio and provides an overlap
check between the endpoint work distributions. Forward and reverse work overlap, endpoint sampling diagnostics, and repeated paths are used to assess each field value. These partition-function
ratios determine the probabilities \(p_s(h)\) in
\eqref{eq:si_crbr3_equilibrium_xi}. The annealing index labels the statistical
bridge and is not physical time.
The longest AIS runs used 736 annealing stages, with four Hamiltonian Monte Carlo updates of 16 leapfrog steps per stage.
A transition bracket is formed by two adjacent simulated field values for which the dominant block-alignment sector changes. The midpoints of the first and last brackets give the field coordinates in main-text Fig.~4F, paired with $n_{\mathrm{first}}^{-1}$ and $n_{\mathrm{last}}^{-1}$, respectively.